%% file: Main.tex
\documentclass[sigconf]{acmart} 

\AtBeginDocument{%
  \providecommand\BibTeX{{%
    \normalfont B\kern-0.5em{\scshape i\kern-0.25em b}\kern-0.8em\TeX}}}
\renewcommand\footnotetextcopyrightpermission[1]{} 
\usepackage{afterpage}
\usepackage[export]{adjustbox}
\usepackage{blindtext}
\usepackage{subcaption}
\usepackage{graphicx}
\usepackage{lipsum}
\usepackage{xspace}
\usepackage{soul,color}
\usepackage{booktabs}
\usepackage{multirow}
\usepackage{array,boldline}
\usepackage{makecell}
\usepackage{algorithm}
\usepackage{algpseudocode}
\usepackage{float}
\usepackage{url}
\hypersetup{
    colorlinks=true,
    linkcolor=blue,
    filecolor=magenta,      
    urlcolor=cyan,
}

\begin{document}

\title{Decamouflage: A Framework to Detect Image-Scaling Attacks on Convolutional Neural Networks}

\renewcommand{\shortauthors}{Trovato and Tobin, et al.}


\author{Bedeuro Kim}
\email{kimbdr@skku.edu}
\affiliation{%
  \institution{Data61, CSIRO, Australia \\ Sungkyunkwan University, South Korea }
  }
  
\author{Alsharif Abuadbba}
\email{sharif.abuadbba@data61.csiro.au}
\orcid{1234-5678-9012}
\affiliation{%
  \institution{Data61, CSIRO, Australia \\ Cyber Security CRC}
}

\author{Yansong Gao}
\email{garrison.gao@data61.csiro.au}
\affiliation{%
  \institution{Data61, CSIRO, Australia \\ Cyber Security CRC}
}

\author{Yifeng Zheng}
\email{Yifeng.Zheng@data61.csiro.au}
\affiliation{%
  \institution{Data61, CSIRO, Australia \\ Cyber Security CRC }}
  
\author{Muhammad Ejaz Ahmed}
\email{Ejaz.Ahmed@data61.csiro.au}
\affiliation{%
  \institution{Data61, CSIRO, Australia }}

\author{Hyoungshick Kim}
\email{hyoung.kim@data61.csiro.au}
\affiliation{%
  \institution{Data61, CSIRO, Australia \\ Sungkyunkwan University, South Korea }}
\author{Surya Nepal}
\email{surya.nepal@data61.csiro.au}
\affiliation{%
  \institution{Data61, CSIRO, Australia \\ Cyber Security CRC}
}

\begin{abstract}
As an essential processing step in computer vision applications, image resizing or scaling, more specifically downsampling, has to be applied before feeding a normally large image into a convolutional neural network (CNN) model because CNN models typically take small fixed-size images as inputs.  However, image scaling functions could be adversarially abused to perform a newly revealed attack called \emph{image-scaling attack}, which can affect a wide range of computer vision applications building upon image-scaling functions.


This work presents an image-scaling attack detection framework, termed as \emph{Decamouflage}. \emph{Decamouflage} consists of three independent detection methods: (1) rescaling, (2) filtering/pooling, and (3) steganalysis. While each of these three methods is efficient standalone, they can work in an ensemble manner not only to improve the detection accuracy but also to harden potential adaptive attacks. \emph{Decamouflage} has a pre-determined detection threshold that is generic. More precisely, as we have validated, the threshold determined from one dataset is also applicable to other different datasets. Extensive experiments show that \emph{Decamouflage} achieves detection accuracy of 99.9\% and 99.8\% in the white-box (with the knowledge of attack algorithms) and the black-box (without the knowledge of attack algorithms) settings, respectively. To corroborate the efficiency of \emph{Decamouflage}, we have also measured its run-time overhead on a {\it personal PC with an i5 CPU} and found that \emph{Decamouflage} can detect image-scaling attacks in milliseconds. Overall, \emph{Decamouflage} can accurately detect image scaling attacks in both white-box and black-box settings with acceptable run-time overhead.

\end{abstract}


\maketitle
\input{Section/1_Introduction.tex}

\input{Section/2_Background.tex}

\input{Section/3_Revealer_Framework.tex}

\input{Section/4_System_Design.tex}

\input{Section/5_Evaluation.tex}

\input{Section/6_Discussion.tex}

\input{Section/7_Related_work.tex}

\input{Section/8_Conclusion.tex}

\bibliographystyle{ACM-Reference-Format}
\bibliography{Reference}

\newpage
\input{Section/9_Appendix}

\end{document}

%% file: Section/1_Introduction.tex

\section{Introduction}
\label{sec:intro}
Deep learning models have shown impressive success in solving various tasks~\cite{krizhevsky2012imagenet,he2016deep,wen2016discriminative,xu2018youtube}. One representative domain is the computer vision that is eventually the impetus for the current deep learning wave~\cite{krizhevsky2012imagenet}. The convolutional neural network (CNN) models are widely used in the vision domain because of its superior performance~\cite{krizhevsky2012imagenet,huang2017densely,he2016deep}. However, it has been shown that deep learning models are vulnerable to various adversarial attacks. Hence, significant research efforts have been directed to defeat the main stream of adversarial attacks such as adversarial samples \cite{szegedy2013intriguing,carlini2017towards}, backdooring \cite{gu2017badnets,liu2017trojaning}, and inference \cite{ganju2018property,lecuyer2019certified}.

Xiao {\it et al.}~\cite{xiao2019seeing} introduced a new attack called \emph{image-scaling attack} (also referred to as \emph{camouflage attack}) that potentially affects all applications using scaling algorithms as an essential pre-processing step, where the attacker's goal is to create attack images presenting a different meaning to humans before and after a scaling operation. This attack would be a serious security concern for computer vision applications. Unlike adversarial examples, this attack is {\it independent} of machine learning models and data. The attack indeed happens before models consume inputs, and hence this type of attack affects a wide range of applications with various machine learning models using image scaling functions. Furthermore, crafted attack images can be used to poison the training data that are typically contributed by third parties or volunteers---a common practice to curate data---that readily enables backdoor attacks when the model is trained over poisoned data (see a detailed example in Section \ref{sec:backdoorAttack}). Herein, the \emph{image-scaling attack} can be used to generate poisoned images bypassing human inspection efficiently because its content and label are consistent visually. Consequently, considering the sequence raised by image-scaling attack, efficient countermeasures are urgently demanded. Below we first give a concise example of the image-scaling attack.

\begin{figure}[!ht]
\centering
\includegraphics[width=0.8\linewidth]{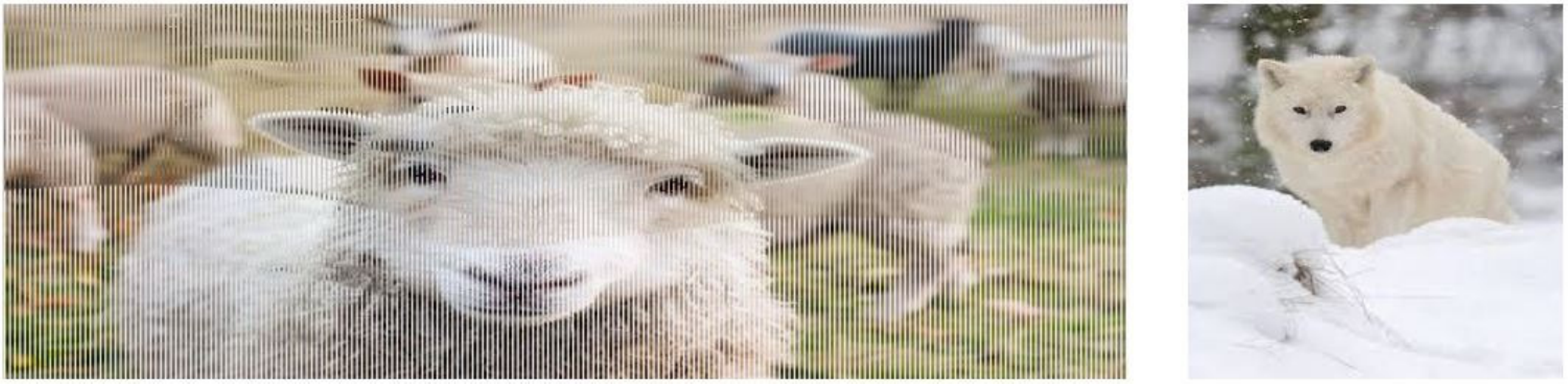}
\caption{Example of image-scaling attacks presenting a deceiving effect. The left image shows what human sees before the scaling operation and the right image shows what the CNN model sees after the scaling operation.}
\label{fig:ImageScaling}
\end{figure}

\textbf{Image-scaling attack example.}
Input of CNN models typically takes fixed-size images such as $224\times 224 \times 3$ (representing the height, width, and the number of color channels) so as to reduce the complexity of computations~\cite{he2016deep}. However, the size of raw input images can be varied or become much larger (e.g., $800 \times 600$) than this fixed-size. Therefore, the resizing or downscaling process is a must before feeding such larger images into an underlying CNN model. Xiao {\it et al.}~\cite{xiao2019seeing} revealed that the image-scaling process is vulnerable to the image-scaling attack, where an attacker intentionally creates an attack image which is visually similar to a base image for humans but recognized as a target image by the CNN model after image-scaling function (e.g., resizing or downscaling) is applied to the attack image. Figure~\ref{fig:ImageScaling} illustrates an example of image-scaling attacks. The `wolf' image is disguised {\it delicately} into the `sheep' image to form an attack image. When the attack image is down-sampled/resized, the `sheep' {\it pixels are discarded}, and the `wolf' image is finally presented. General, image-scaling attack abuses an inconsistent understanding of the same image between humans and machines. 



The strength of the image-scaling attack is its independence on CNN models and data --- it requires no knowledge of training data and the model because it mainly exploits the image scaling function used for pre-processing. For image-scaling attacks, only the knowledge about the used image-scaling function is required. It is noted that the attacker can relatively easily obtain this information because a small number of well-known image scaling functions (e.g., nearest-neighbor, bilinear, and bicubic interpolation methods) are commonly used for real-world services, and a small number of input sizes (e.g., $224\times224$ and $32\times32$) are used for representative CNN models~\cite{xiao2019seeing}, as shown in Table~\ref{tb:inputsize}. Furthermore, the parameters for the image-scaling function can be exposed to the public in some services. Nonetheless, even when the parameter information is not provided explicitly, it is feasible to infer the function parameter information used in a target service with API queries under a limited trial by an attacker~\cite{xiao2019seeing}.

\begin{table}[!ht]

\caption{Input sizes for popular cnn models.}
\scalebox{0.85}{
\begin{tabular}{>{\centering\arraybackslash}m{2in} >{\centering\arraybackslash}m{1in}}

\hlineB{1.5}
Model   & Size \\
        &  (pixels\,*\,pixels) \\ \hline
LeNet-5                           & 32\,*\,32                                                          \\ \hline
VGG, ResNet, GoogleNet, MobileNet & 224\,*\,224                                                        \\ \hline
AlexNet                           & 227\,*\,227                                                        \\ \hline
Inception V3/V4                   & 299\,*\,299                                                        \\ \hline
DAVE-2 Self-Driving               & 200\,*\,66                                                         \\ 
\hlineB{1.5}
\end{tabular}}
\label{tb:inputsize}
\end{table}

The image-scaling attacks can target various surfaces. First, as an evasive attack, the attack images crafted via image-scaling attacks can achieve the attack effect similar to adversarial examples with an advantage of agnostic to underlying CNN models. Second, the attack image can be exploited for data poisoning to insert a backdoor into {\it any} model trained over the poisonous data (see Section \ref{sec:backdoorAttack}). 

Unlike other adversarial attacks where corresponding countermeasures have been well investigated, only one study suggested defense mechanisms against image scaling attacks. Quiring {\it et al.} \cite{quiring2020adversarial} first analyzed the root cause of image scaling attacks and proposed two defense mechanisms, (1) use of robust scaling algorithms and (2) image reconstruction, to prevent image-scaling attacks by delicately exploiting the relationship between the downsampling frequency and the convolution kernel used for smoothing pixels. The proposed defense mechanism sanitizes those pixels, which renders the image-scaling attack technique unable to inject target pixels with the required quality. However, their defense approaches have the following downsides. First, the use of robust scaling algorithms is likely to cause backward compatibility problems with existing scaling algorithms in OpenCV and TensorFlow. Moreover, as Quiring {\it et al.}~\cite{quiring2020adversarial} mentioned, small artifacts from an attack image can remain even after applying their suggested scaling algorithms, as the manipulated pixels are not cleansed and still contribute to the scaling. Second, the image reconstruction method removes the set of pixels in the attack images and reconstructs those pixels with image filters. This approach would significantly decrease the attack chance, but it can inherently degrade the quality of input images for CNN models.

To obviate image quality degradation and potential incompatibility with {\it prevention} mechanisms, we focused on developing a solution to {\it detect} attack images regarding the image-scaling attack, including one novel angle e.g., treating the image-scaling attack as a kind of steganography for information hiding. We aim to develop a defense mechanism to detect attack images only without any modifications to input images for CNN models. Also, we develop \textit{Decamouflage} as an independent module compatible with any existing scaling algorithms---alike a plug-in protector. Furthermore, \textit{Decamouflage} is designed for detecting attack images crafted via image-scaling attacks even under black-box settings where there is no prior information about the attack algorithm.




Our key contributions are summarized as follows:

\begin{itemize}
    \item \textit{Decamouflage} is the first practical solution to detect image-scaling attacks. We develop three different detection methods (scaling, filtering, and steganalysis) and construct \textit{Decamouflage} as an ensemble of those methods. Each method can be deployed individually and eventually work together as complementary to each other to maximize the detection accuracy. Our source code is released at \url{https://github.com/anynymous/Decamouflage}\footnote{The artifacts including source code will be released upon the publication.}.
    \item We identify three fundamental metrics (mean squared errors (MSE), structural similarity index (SSIM), and centered spectrum points (CSP)) that can be used to distinguish benign images from attack images generated by image-scaling attacks. Those metrics would also be  applicable for continuous research in the line of detecting attack images.
    \item We empirically validate the feasibility of \textit{Decamouflage} for both the white-box setting (with the knowledge of the attacker's algorithm) and the black-box setting (without the knowledge of the attacker's algorithm). We demonstrate that \textit{Decamouflage} can be effective in both settings with experimental results. 
    \item We evaluate the detection performance of \textit{Decamouflage} using an unseen testing dataset to show its practicality. We used the ``NeurIPS 2017 Adversarial Attacks and Defences Competition Track'' image dataset~\cite{kurakin2018adversarial} to find the optimal thresholds for \textit{Decamouflage} and used the ``Caltech 256'' image dataset~\cite{Caltech_image} for testing. To implement image-scaling attacks, we use the code released in the original work by Xiao {\it et al.}~\cite{xiao2019seeing}. The experimental results demonstrate that \textit{Decamouflage} achieves detection accuracy of 99.9\% with a false acceptance rate of 0.2\% and a false rejection rate of 0.0\% in the white-box setting, and detection accuracy of 99.8\% with a false acceptance rate of 0.3\% and a false rejection rate of 0.1\% even in the black-box setting. In addition, the run-time overhead of \textit{Decamouflage} is less than 174 milliseconds on average evaluated with a {\it personal PC} with an Intel Core i5-7500 CPU (3.41GHz) and 8GB memory, indicating that \textit{Decamouflage} can be deployed for online detection. 
\end{itemize}

%% file: Section/2_Background.tex
\section{Background}
\label{sec:Bg}

In this section, we provide the prior knowledge for the image-scaling attack and its enabled insidious backdoor attack.

\subsection{Image-Scaling Attack}
The preprocessing steps for input images in a typical deep learning pipeline is an essential stage. Recently, Xiao {\it et al.}~\cite{xiao2019seeing} demonstrated a practical adversarial attack targeting the scaling functions used by widely used deep learning frameworks. The attack exploited the fact that deep learning-based models accept only small fixed-size input images. As presented in Table \ref{tb:inputsize}, nine popular deep learning models are summarized, and they all use a fixed input scale during both training and inference phases. In practice, images are often captured on larger dimensions than what models expect; therefore, downscaling operations are necessary for such situations. Thus an adversary has the chance to modify an image to adversarially change its content seen by the model after undergoing downscaling.  

\begin{figure}[!h]
\centering
\includegraphics[width=0.8\linewidth]{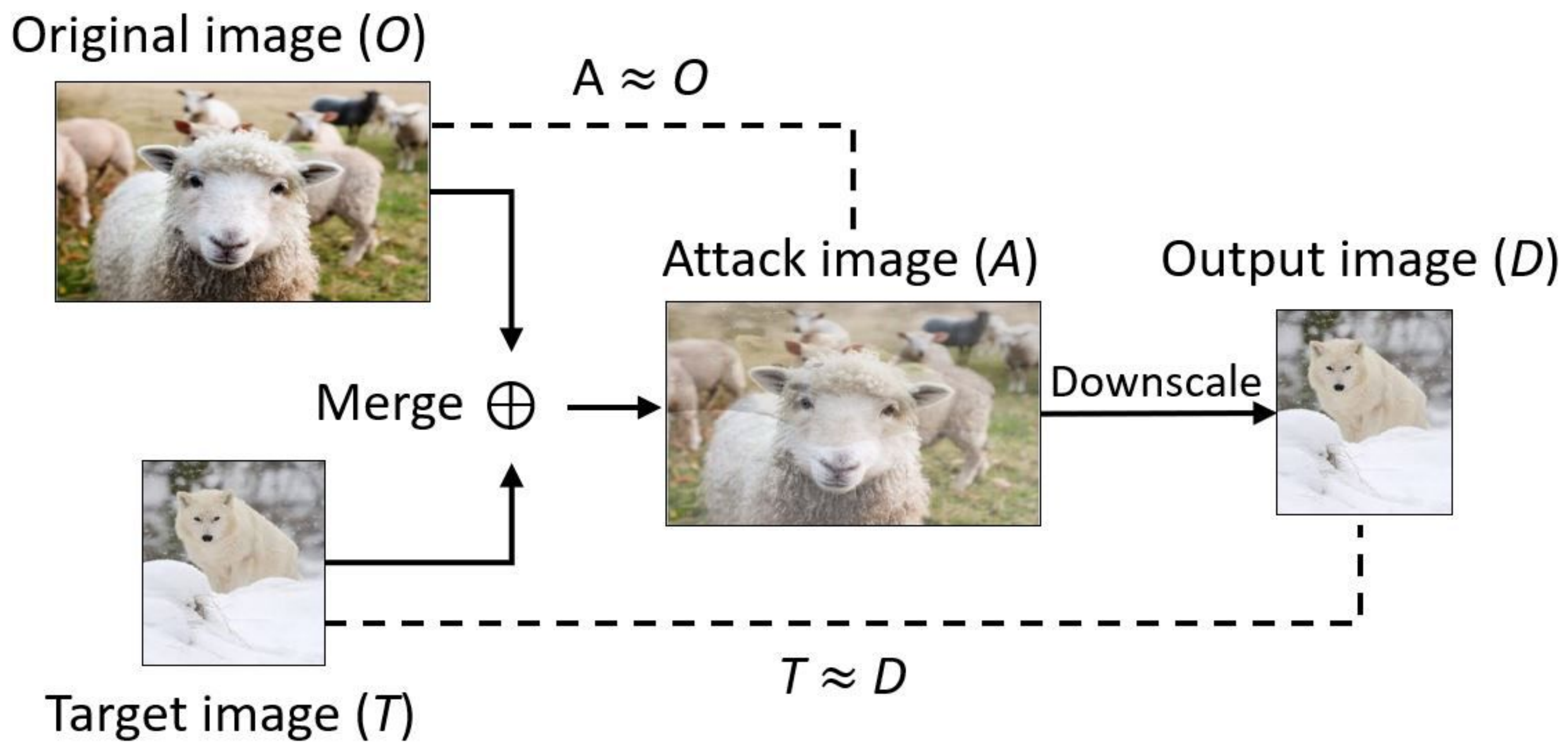}
\caption{Overall process of an image-scaling attack. An adversary creates an attack image $A$ (tampered sheep image) such that it looks like $O$ (original sheep image) to humans, but it is recognized as $T$ (targeted wolf image) by CNN models after applying image scaling operations. Here $X \approx Y$ represents that $X$ looks similar to $Y$.}
\label{fig:crafed image}
\end{figure}

One example is illustrated in Figure~\ref{fig:crafed image}, where a wolf is disguised into a sheep image. The human sees sheep, but the model sees a wolf once the tampered sheep image undergoes the downsampling step. More precisely, the adversary slightly alters an original image $O$ so that the obtained attack image $A = O + \Delta$ resembles a target image $T$ once downscaled. The attack mechanism can be demonstrated as the following quadratic optimization problem:

\vspace{0.3mm}
\begin{equation}
    min(||\Delta ||_{2}^{2})  \; \; s.t. \; \; ||scale(O + \Delta ) - T||_{\infty} \leq \epsilon 
\end{equation}
\vspace{0.3mm}

Also, each pixel value of $A$ needs to be maintained within the fixed range (e.g., [0,255] for 8-bit images). This problem can be solved with Quadratic Programming (QP)~\cite{gu2017badnets}. The successful attack criteria are that the obtained image $A$ should be visually similar to the original image $O$, but the downscaled output $D$ should be recognized as the target image $T$ after scaling. In other words, the attack has to satisfy two properties:

\begin{itemize}
    \item The resultant attack image $A$ should be visually indistinguishable from the original image $O$ ($A \approx O$).
    \item The output image $D$ downscaled from the attack image $A$ should be recognized as the target image $T$ by CNN models ($T \approx D$).
\end{itemize}

\subsection{Image-Scaling Attack Assisted Backdooring}
\label{sec:backdoorAttack}
The image-scaling attack greatly facilities backdoor attack that is one emerging security threat to current ML pipeline. The backdoored model behaves the same to its counterpart, the clean model, in the absence of the trigger~\cite{gao2020backdoor}. However, the backdoored model is hijacked to misclassify any input with the trigger to the attacker's target label. This newly revealed backdoor attack does need to tamper the model to insert the backdoor first. The attack surface of the backdoor is regarded wide: data poisoning is among one main attack surface~\cite{gao2020backdoor}. In this context, the user collects data from many sources, e.g., public or contributed by volunteers or third parties. Since the data sources could be malicious or compromised, the curated data could be poisoned. Image-scaling attack facilitates data poisoning attack to insert a backdoor into the CNN model~\cite{gao2020backdoor}, which was already demonstrated explicitly by Quiring {\it et al.}~\cite{quiring2020backdooring}. 

Here, we exemplify this backdoor attack using face recognition. First, the attacker randomly selects a number of images from different persons, e.g., Alice, Bob. The attacker also chooses black-frame eye-glass as the backdoor trigger. Second, the attacker poisons both Alice and Bob face images by stamping the trigger---these poisonous images afterward referred to as trigger images. Third, assisted with an image-scaling attack, the attacker disguises the trigger image into administer's image---this means the targeted person of the backdoor attack is the administer. A number of attack/poisoned images are crafted and submitted to the data aggregator/user. As the attack image's content is consistent with its label -- the attack image still visually indistinguishable from the administer's face, the data aggregator cannot identify the attack image. Fourthly, the user trains a CNN model over the collected data. In this context, the attack images seen by the model are trigger images. Therefore, the CNN model is backdoored, which learns a sub-task that associates the trigger with the administer. During the inference phase, when any person, e.g., Eve, wears the black-frame eye-glass indicating a trigger, the face recognition system will misclassify Eve into the administer.

%% file: Section/3_Revealer_Framework.tex
\section{Potential Detection Methods: Key Insights}
\label{sec: Potential Detection Methods}

To proactively defeat the image-scaling attack, one would first identify potential methods from different angles. Therefore, the first research question (RQ) is as below.   

\begin{center}
    \noindent{\textit{\textbf{\,RQ. 1: What are the potential methods to reveal the target image embedded by the image-scaling attack?\qquad}}}
\end{center}

This work identifies three efficient methods and visualizes their ability to detect that attack. Here we provide a general concept for each method. We exchangeably use the terms original image and benign image in the rest of this paper.

\subsection{Method 1: Scaling Detection}
\label{sec: Scaling Detection}

\begin{figure}[!ht]
  \centering
  \begin{subfigure}[r]{0.5\textwidth}
    \includegraphics[trim=0 0 0 0, width=0.8\linewidth]{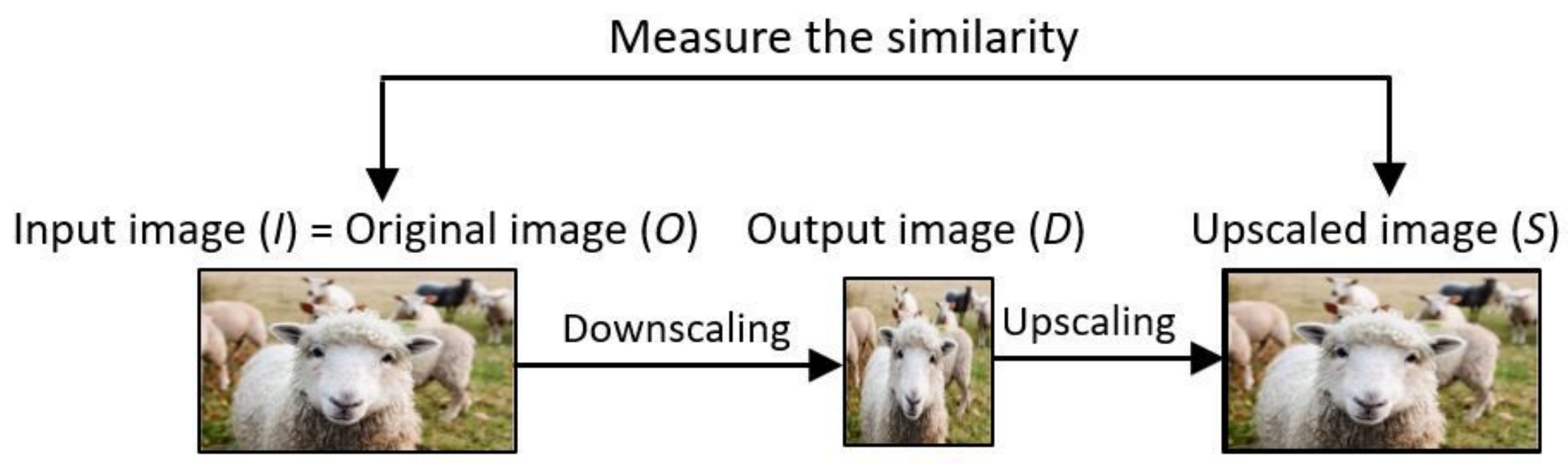}
    \caption{Benign case.}
  \end{subfigure}
  \vspace{2mm}
   
   \begin{subfigure}[r]{0.5\textwidth}
    \includegraphics[trim=0 0 50 0,width=0.90\linewidth ]{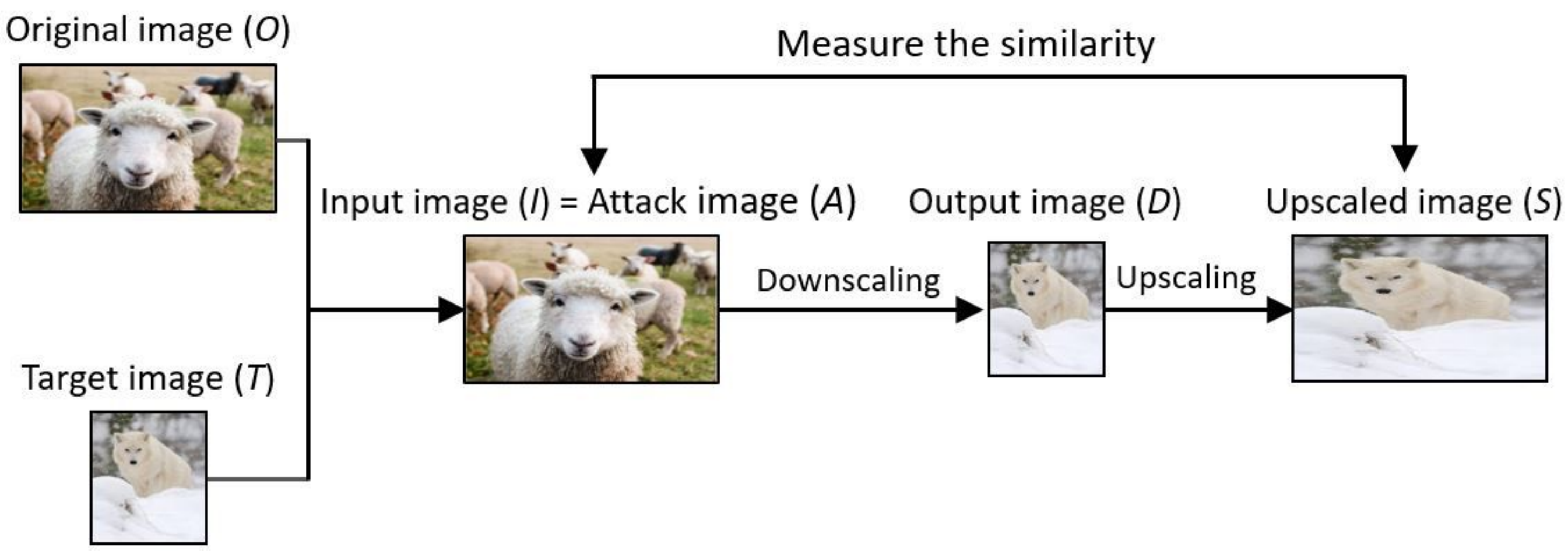}
    \caption{Attack case.}
  \end{subfigure}
\caption{Overview of the scaling detection method. We obtained the upscaled image $S$ from the downscaled image $D$ and then measured the image similarity between $S$ and the input image $I$. If the input image $I$ is a benign image (i.e., original image $O$), $S$ will remain similar to $I$; otherwise, $S$ would be significantly different from $I$.
}
\label{fig:scaling_detection}
\end{figure}

We first explore the potential of reverse-engineering the attack process. In the attack process, the attack image $A$ is downsampled to the output image $D$ to be recognized as $T$ for CNN models. Therefore, we need to upscale the output image $D$ to the upscaled image $S$ in the reverse engineering process. Based on the reverse engineering process, we design an image-scaling attack detection method as follows. Given an input image $I$ (which can potentially be an attack image) for a CNN model, we apply the downscaling and upscaling operations in sequence to obtain the image $S$ and measure the similarity between $I$ and $S$. Our intuition is that if the input image $I$ is a
benign image (i.e., the original image $O$), $S$ will remain similar to $I$; otherwise, $S$ would be significantly different from $I$ (see Figure \ref{fig:scaling_detection}).

Xiao {\it et al.}~\cite{xiao2019seeing} suggested the color histogram as an image similarity metric for detecting attack images without conducting experiments. However, we found that the color histogram is not a valid metric for the purpose of detecting image-scaling attacks. Our observation is consistent with the results in \cite{quiring2020backdooring}. Therefore, it is challenging to find a proper metric to distinguish the case of attack images from benign images. We will discuss this issue in Section~\ref{sec:systemDesign}.

\subsection{Method 2: Filtering Detection}
\label{sec: Filtering Detection}
\begin{figure}[!h]
\centering
\includegraphics[width=0.7\linewidth]{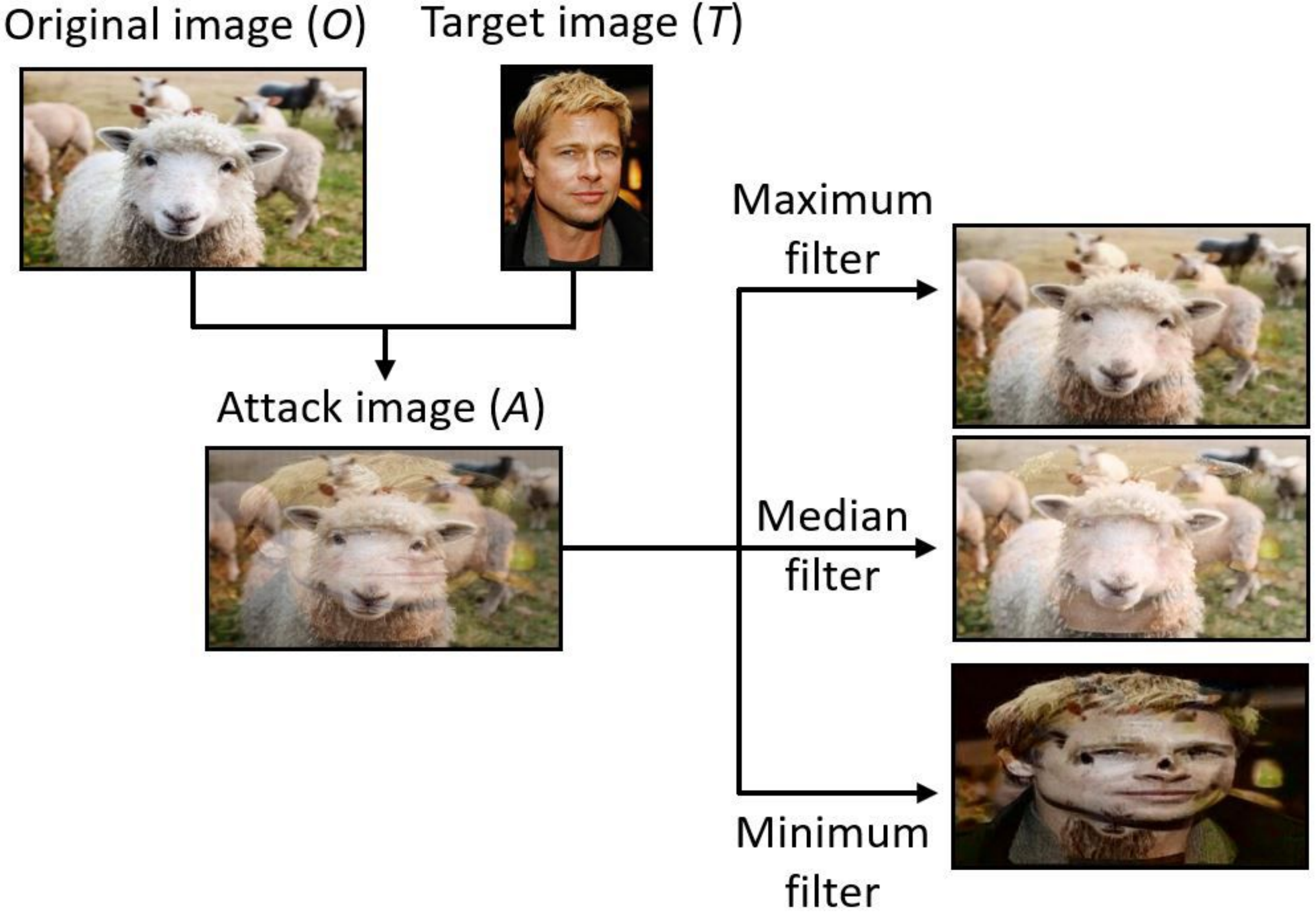}
\caption{Image filter results on an attack image.}
\label{fig:min_max detection}
\end{figure}

The image-scaling attack relies on embedding the target image pixels within the original image pixels to avoid human visual inspection by abusing image scaling functions. Therefore, if we use image filters to remove noises, the embedded target image pixels might be removed or affected because the embedded target image pixels would be significantly different from the original image pixels. Figure~\ref{fig:min_max detection} shows the results of an attack image after applying the minimum filter~\cite{schalkoff1989digital}, the median filter, and the maximum filter, respectively.\footnote{We used the OpenCV image filtering APIs (see \url{https://docs.opencv.org/2.4/modules/imgproc/doc/ filtering.html}).} We can see that 
the minimum filter reveals the target image compared with the other filters. 

Based on this observation, we suggest another image-scaling attack detection method. Given an input image $I$ (which can potentially be an attack image) for a CNN model, we apply an image filter to obtain the image $F$ and measure the similarity between $I$ and $F$. Our intuition is that if the input image $I$ is a benign image (i.e., the original image $O$), $F$ will remain similar to $I$; otherwise, $F$ would be significantly different from $I$. For this purpose, we specifically select the minimum filter because it could effectively remove the original image pixels in the case of attack images. 

\begin{figure}[!h]
\centering
\includegraphics[trim=0 0 100 0,width=0.6\linewidth]{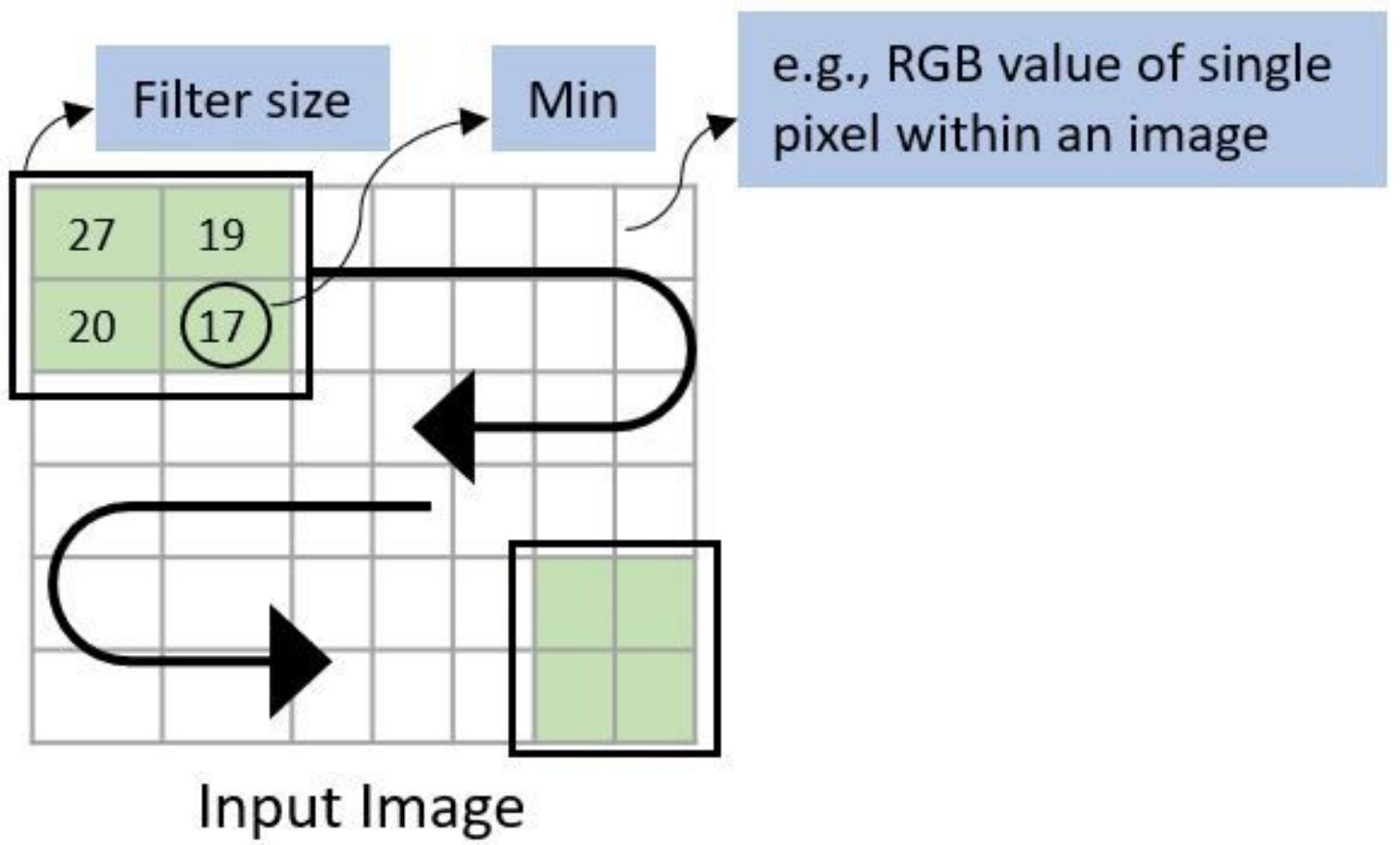}
\caption{Process of applying the minimum filter.}
\label{fig:filter_shape}
\end{figure}

The minimum filter is used with fixed window size. Figure~\ref{fig:min_max detection} illustrates how the minimum filter works on an image. The image filtering process is done by dividing the image $M\times N$ into smaller 2D blocks $x_{i=1} ^{b}\times y_{j=1} ^{b}$ where $b$ is the number of blocks and $x,y$ are the filter size. If we use the $2 \times 2$ minimum filter, only the smallest pixel value among a neighborhood of the block $x_{i} \times y_{j}$ is selected as shown in Figure \ref{fig:filter_shape}. For applying the minimum filter, the smallest pixel value from each block is selected.

We will discuss how to measure the image similarity between $I$ and $F$ and determine whether a given image is an attack image in Section \ref{sec:systemDesign}.




\subsection{Method 3: Steganalysis Detection}
\label{sec:sd}

The image-scaling attack's key idea is to embed the target image as cluttered pixels so that they are less recognized by human eye perceptuality. Consequently, we treat the perturbed pixels as information that the attacker tries to hide in this method, which is similar to steganography ~\cite{shih2017digital}. Steganography is a technique of hiding information in digital media such as images to avoid secret data detection by unintended recipients. Therefore, we may constructively employ steganalysis mechanisms to expose the hidden perturbed pixels embedded by the image-scaling attack based on the similarity between the image-scaling attack and steganography.

We explore the frequency domain based steganalysis mechanism to find out the perturbed pixels within the attack image. Fourier Transform (FT) is an operation that transforms data from the time (or spatial) domain into the frequency domain \cite{van2018signal}. Because an image consists of discrete pixels rather than continuous patterns, we use the Discrete Fourier Transformation (DFT)~\cite{bracewell1986fourier}. We first transform the input (potential attack) image $A$ into the 2-dimensional space, namely spectrum image. For a square image of size $N\times N$, the 2-dimensional DFT is given by:

\begin{equation}
 F(k,l)=\sum_{i=0}^{N-1}\sum_{j=0}^{N-1}f(i,j))e^{-i2\pi (\frac{k_i}{N}+\frac{l_i}{N})}   
\end{equation}
\vspace{0.3mm}

where $f(i,j)$ is the spatial domain images, and the exponential term is the corresponding basis function to each $F(k,l)$ point in the DFT space. The basis functions are sine and cosine waves with increasing frequencies as depicted below:

\begin{equation}
    \left [  cos \left ( {2\pi (\frac{k_i}{N}+\frac{l_i}{N})} \right ) - i \cdot sin  \left ( {2\pi (\frac{k_i}{N}+\frac{l_i}{N})} \right ) \right]
\end{equation}
\vspace{0.3mm}

The resultant DFT spectrum contains the low and high-frequency coefficients. The low frequencies capture the image's core features, whereas the high frequency reflects the less significant regions within an image. Direct visualization of both frequencies shows that a broad dark region in the middle represents the high frequency, while low frequency appears as a whiter clattered area on the edges. This visualization can not provide us with an automated quantification to distinguish attack images from benign images. Therefore, we apply logarithmic with a shift to flip the whiter frequency to centralize the low frequencies called centered spectrum as given by: 

\begin{equation}\label{eq:shiftlog}
    F(x,y)=\sum_{k=0}^{N-1}\sum_{l=0}^{N-1} \log  \left |  \Theta \cdot F(k,l) \right | 
\end{equation}
\vspace{0.3mm}

where $\Theta$ is the predetermined shift for $F(k,l)$ low-frequency point. 

\begin{figure}[!ht]
\centering
\includegraphics[width=0.7\linewidth]{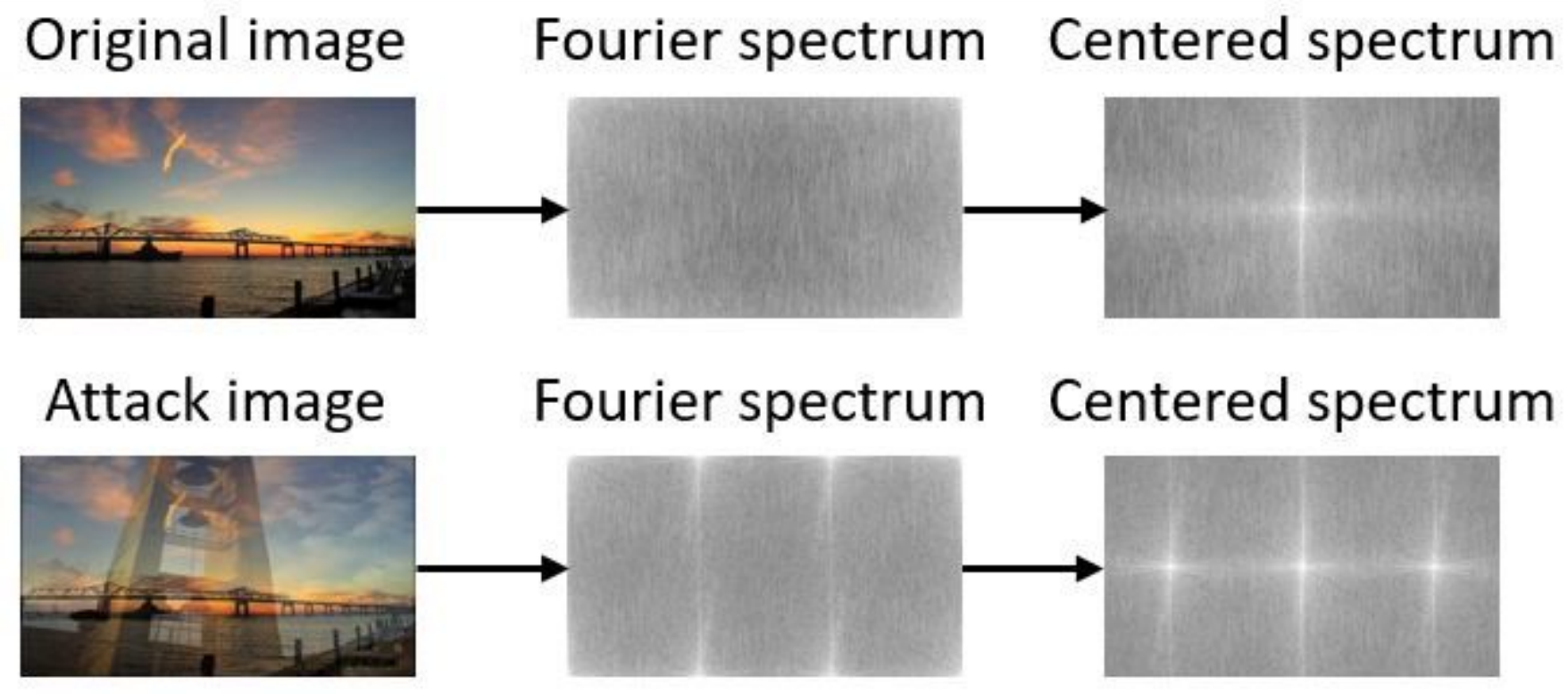}
\caption{Results of centered spectrum points on a benign image and an attack image.}
\label{fig:stega_detection}
\end{figure}

If we apply the FT operation on a benign image, a benign image has one centered spectrum point. However, as shown in Figure~\ref{fig:stega_detection}, attack images overall exhibit multiple centered spectra as opposite to one centered spectrum point observed in benign images because the cohesion of the original image pixels is broken due to the arbitrary perturbation to embed the target image pixels. 

Based on this observation, we suggest an image-scaling attack detection method using the frequency domain based steganalysis. Given an input image $I$ (which can potentially be an attack image) for a CNN model, we convert it into a Fourier spectrum to obtain the image $B$ and count the centered spectrum points in $B$. We will discuss how to count the number of the centered spectrum points and determine whether a given image is an attack image in Section \ref{sec:systemDesign}.


\noindent{\bf Summary:} As an answer to RQ. 1, we suggest that three detection methods (scaling, filtering, and steganalysis) can potentially expose attack images generated by image-scaling attacks. Each method is designed based on a different insight/angle to detect image-scaling attacks. The scaling detection and filtering detection methods are designed to detect the image-scaling attacks in the spatial domain, while the steganalysis method is designed to detect the image-scaling attacks in the frequency domain.

%% file: Section/4_System_Design.tex
\section{Decamouflage System Design}\label{sec:systemDesign}
In this section, we provide the Decamouflage framework exploiting the above-identified detection methods to answer the RQ. 2: 

\vspace{0.06mm}
\begin{center}
    \noindent{\textit{\textbf{RQ. 2: How can we develop an automated process to detect image-scaling attacks using the identified methods?}}}
\end{center}
\vspace{0.08mm}

We first define the threat models that we focused on in this paper. Next, we introduce three key metrics to find image-scaling attacks in an automated manner. We finally provide an overview of the \textit{Decamouflage} detection system that can efficiently distinguish attack images from benign images with the methods identified in Section~\ref{sec: Potential Detection Methods}.

\subsection{Threat Model}\label{section:TM4.2}
For a defense mechanism, we consider both white-box and black-box settings. In the white-box setting, we assume that the defender (i.e., service provider) knows the attacker's algorithm; thus, the parameters for \textit{Decamouflage} are determined to target for the attacker's specific algorithm. In the black-box setting, we assume that the defender does not know the attacker's algorithm. Perhaps, the black-box setting seems more practical because it would be difficult to obtain information about the attacker's algorithm, and we should also consider many different conditions for the image-scaling attack.

\textit{Decamouflage} can be performed offline and online. Offline is suitable for defeating backdoor attack assisted with image-scaling attack (presented in Section \ref{sec:backdoorAttack}). Herein, the defender is the data aggregator/user who has access to attack images. In this case, we reasonably assume that the user owns a small set, e.g., 1000 of hold-out samples produced in-house. The defender must remove attack images crafted by image-scaling attacks to avoid backdoor insertion in the trained model. On the other hand, for online detection, \textit{Decamouflage} is to tell whether input images are attack images or benign images during run-time. 

\subsection{Metrics for Decamouflage}\label{sec:residual}

\textit{Decamouflage} is basically built as an ensemble solution on the three image-scaling attack detection approaches presented in Section~\ref{sec: Potential Detection Methods}. Therefore, it is essential to quantify the differences between attack images and benign images for each approach.

\begin{figure}[!h]
\centering
\includegraphics[width=0.9\linewidth]{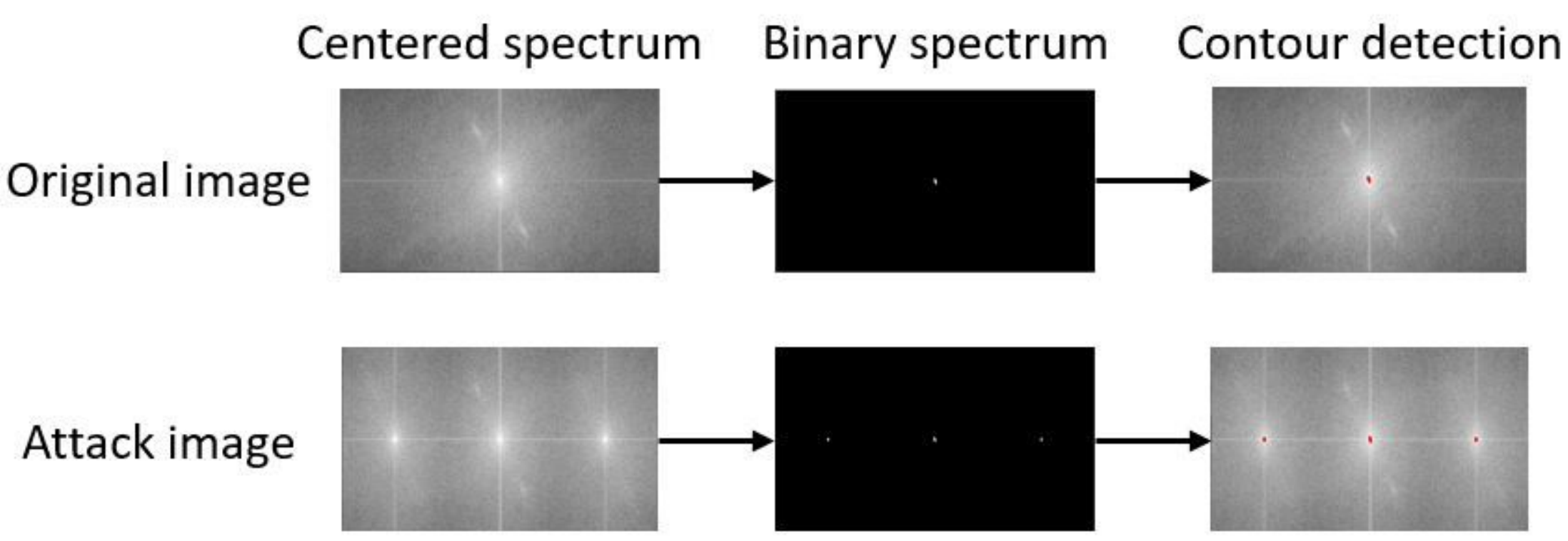}
\caption{Process of computing the centered spectrum points on an original image and an attack image. Given an image, we first apply the FT operation and then apply a low pass filter to extract the low frequencies of the image only (see `Binary spectrum'). Finally, we count the number of centered spectrum points using a contour detection algorithm. In this example, we can see three centered spectrum points in the attack image while there is only one centered spectrum point in the original image.}
\label{fig:contourDetect}
\end{figure}

Here, we recommend using MSE and SSIM~\cite{hore2010image} for scaling detection~\ref{sec: Scaling Detection} and filtering detection~\ref{sec: Filtering Detection} methods. We considered several metrics such as peak signal-to-noise ratio (PSNR) (see Appendix~\ref{sec: PSNR ineffective as a threshold}) but we found that MSE and SSIM are most suitable for \textit{Decamouflage}. As for the steganalysis detection method~\ref{sec:sd}, we recommend using the number of centered spectrum points. The definition of each metric is as follows:

\begin{itemize}
    \item \textbf{MSE} computes the average of the squares of the differences between two images $A$ and $B$ as given in Equation.\ref{eq:mse}, where $y_i$ is the $i$th pixel in the image $A$; $\widetilde{y_i}$ is the $j$th pixel in the image $B$; and $n$ is the size of $A$\footnote{In \textit{Decamouflage}, we use the same size of input images $A$ and $B$.}.

\begin{equation}\label{eq:mse}
    MSE = \frac{1}{n}\sum_{i=1}^{n}\left ( y_1 - \widetilde{y_i} \right )^2
\end{equation}
\vspace{0.3mm}

    \item \textbf{SSIM} index is another popularly used metric to compute the similarities of local luminance, contrast, and structure between two images due to its excellent performance and simple calculation. The SSIM index can be calculated in windows with different sizes (block unit or image unit) for two images. The SSIM index between two images $A$ and $B$ can be calculated as follows:

\begin{equation}
    SSIM(A, B)=\frac{\left ( 2\mu_A\mu_B+c_1 \right )\left ( 2\sigma_{AB} +c_2 \right )}{\left ( \mu^2_A+\mu^2_B+c_1 \right )\left ( \sigma^2_A+\sigma^2_B+c_2 \right )}
\end{equation}
\vspace{0.3mm}
where $\mu_A\mu_B$ are the average of $A$ and $B$; $\sigma^2_A+\sigma^2_B$ and $\sigma_{AB} $ are their variance and covariance, respectively. Here, $c_1$ and $c_2$ are variables to stabilize the division with weak denominator.

    \item \textbf{CSP} is the number of centered spectrum points on an image in the frequency domain space. To count this number from a given image, we first apply the FT operation and then apply a low pass filter to allow only low frequencies. Given a radius value $D_T$ as a threshold, our low pass filter can be modeled as follows:

\begin{equation}
    H(u,v)=\left\{\begin{matrix}
 1& if \;D(u,v)\leq D_T\\ 
 0& if \;D(u,v)>  D_T
\end{matrix}\right.
\end{equation}
\vspace{0.3mm}
    Finally, after applying the low pass filter on the image, we obtain a binary spectrum image containing low frequencies only. The number of bright low-frequency points is then automatically counted by using a contour detection function. This process is visualized in Figure~\ref{fig:contourDetect}.
\end{itemize}

\subsection{Overview of Decamouflage}

\begin{figure}[!ht]
\centering
\includegraphics[width=1.\linewidth]{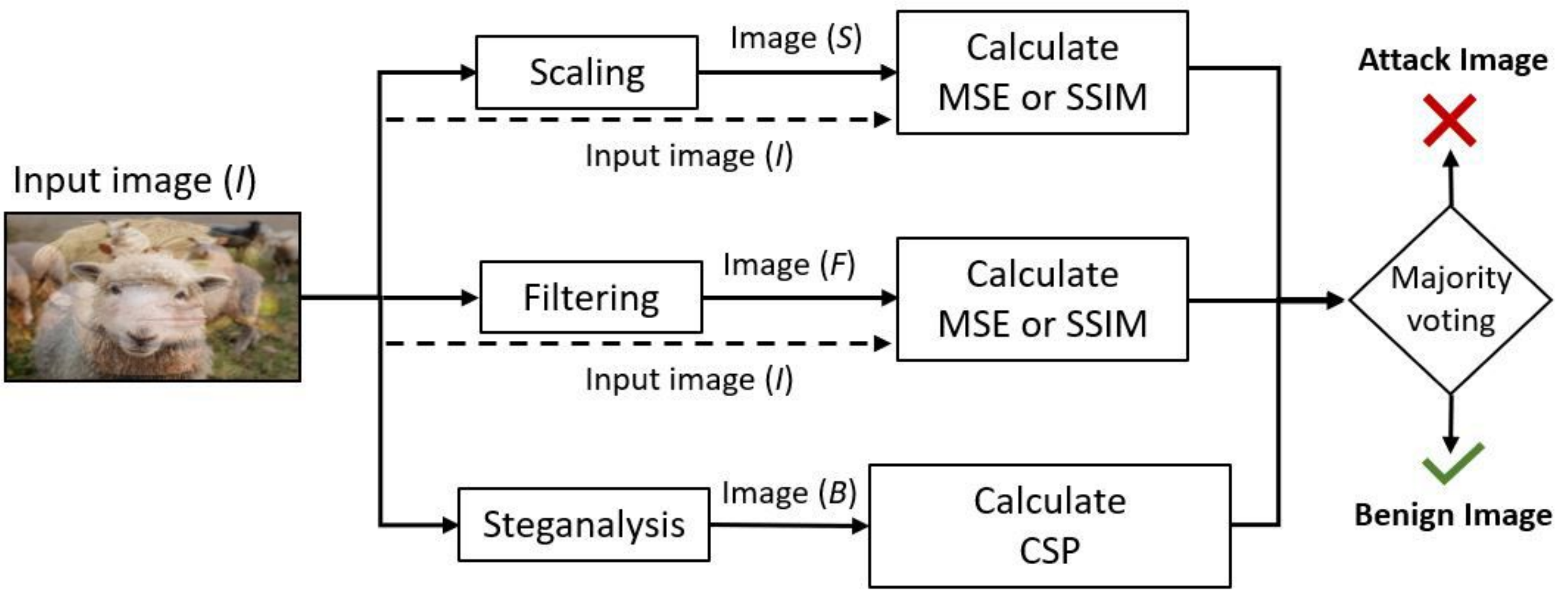}
\caption{Overview of Decamouflage.}
\label{fig:system design}
\end{figure}

The overview of \textit{Decamouflage} is illustrated in Figure ~\ref{fig:system design}, whereas each of the three methods is detailed in Algorithm \ref{alg:scaling},\ref{alg:filtering} and \ref{alg:steganalysis}, respectively.  Given an input image $I$ (which can potentially be an attack image) for a CNN model, \textit{Decamouflage} runs the three methods (described in Algorithm \ref{alg:scaling}, \ref{alg:filtering} and \ref{alg:steganalysis}) yielding the decision individually in parallel, and then performs majority voting (ensemble technique) to determine whether $I$ is an attack image crafted by the image-scaling attack or not.


Algorithm \ref{alg:scaling} describes the computational procedure of the scaling detection method. In this algorithm, we initially set $Attackflag$ to $False$ (line 3). We convert the input image $I$ into $D$ using a downscaling operation and then convert $D$ into $S$ using an upscaling operation (lines 4--5). Next, we calculate either ${MSE}_{(I, S)}$ or $SSIM_{(I, S)}$ between $I$ and $S$ depending on $Metricflag$ indicating which metric is used (line 6--12). If the calculated metric value $Score$ is greater than or equal to the predefined threshold $Score_T$, we set $Attackflag$ to $False$ (lines 13--15). Similarly, we design Algorithm \ref{alg:filtering} and \ref{alg:steganalysis}, but we skip the details of those algorithms from this paper due to the paper page limit.

To use each method effectively, we empirically set the threshold value for the method. Our recommended threshold values are presented in Section~\ref{sec: Experimental Setup}.  



\begin{algorithm}[h]
\caption{Scaling detection}
\label{alg:scaling}
\begin{algorithmic}[1]
\Procedure {Scaling detection}{$I$, $Metricflag$}   \\\Comment{$I$: input image, $Metricflag$: input metric flag}   
    \State $Attack\,flag \leftarrow False$ 
    \State \texttt{$D \leftarrow scale\,down(I)$}  \Comment{$D$: downscaled image}  
    \State \texttt{$S \leftarrow scale\,up(D)$} \Comment{$S$: upscaled image}
    \If {$Metricflag == True$}
        \State $Score \leftarrow {MSE}_{(I, S)}$
        \State $Score_{T} \leftarrow {MSE}_{T}$
        \Comment{${MSE}_{T}$: MSE Threshold}
    \Else
        \State $Score \leftarrow {SSIM}_{(I, S)}$
        \State $Score_{T} \leftarrow {SSIM}_{T}$
        \Comment{${SSIM}_{T}$: SSIM Threshold}
    \EndIf
    \If {$Score \geq {Score}_{T}$} 
        \State $Attack\,flag \leftarrow True$
    \EndIf
    \State \Return $Attack\,flag$
\EndProcedure
\end{algorithmic}
\end{algorithm}

\begin{algorithm}[h]
\caption{Filtering detection}
\label{alg:filtering}
\begin{algorithmic}[1]

\Procedure {Filtering detection}{$I$, $Metricflag$}    \\\Comment{$I$: input image, $Metricflag$: input metric flag}  
    \State $Attack\,flag \leftarrow False$
        \State \texttt{$F \leftarrow minimum\,filter(I)$} \Comment{$F$: filtered image} 
        
    \If {$Metricflag == True$}
        \State $Score \leftarrow {MSE}_{(I, F)}$
        \State $Score_{T} \leftarrow {MSE}_{T}$
        \Comment{${MSE}_{T}$: MSE Threshold}
    \Else
        \State $Score \leftarrow {SSIM}_{(I, F)}$
        \State $Score_{T} \leftarrow {SSIM}_{T}$
        \Comment{${SSIM}_{T}$: SSIM Threshold}
    \EndIf
        
   \If {$Score \geq {Score}_{T}$} 
        \State $Attack\,flag \leftarrow True$
    \EndIf
    \State \Return $Attack\,flag$
\EndProcedure

\end{algorithmic}
\end{algorithm}

\begin{algorithm}[h]\label{alg:steganalysis}
\caption{Steganalysis detection}
\label{alg:steganalysis}
\begin{algorithmic}[1]

\Procedure {Steganalysis detection}{$I$}   \Comment{$I$: input image}     
    \State $Attack\,flag \leftarrow False$ 
        \State \texttt{$C \leftarrow centered\,spectrum\,image(I)$} 
        \\\Comment{$C$: centered spectrum image} 
        \State \texttt{$B \leftarrow convert\,binary(C)$} \Comment{$B$: binary image}
        \State \texttt{${CSP}_{B} \leftarrow Count\,the\,centered\,spectrum\,points\,in\,B$} \\\Comment{${CSP}_{B}$: the number of centered spectrum points in $B$}
    
    \If{${CSP}_{B} \geq {CSP}_{T}$}
        \Comment{${CSP}_{T}$: CSP threshold} 
        \State $Attack\,flag \leftarrow True$
    \EndIf
    \State \Return $Attack\,flag$
\EndProcedure

\end{algorithmic}
\end{algorithm}

\noindent{\bf Summary:} As an answer to RQ. 2, we present \textit{Decamouflage} to detect image-scaling attacks in an automated manner. To achieve this goal, we suggest three metrics (MSE, SSIM, and CSP) that can be effectively used for the three techniques in Section~\ref{sec: Potential Detection Methods}.

%% file: Section/5_Evaluation.tex
\section{Evaluation}\label{sec:evaluation}
This section introduces the experiment setup and performance evaluation for \textit{Decamouflage}.

\subsection{Experiment Setup}
\label{sec: Experimental Setup}

For a more practical testing environment, we consider evaluating the performance of \textit{Decamouflage} for an unseen dataset. We used ``NeurIPS 2017 Adversarial Attacks and Defences Competition Track''~\cite{kurakin2018adversarial} to select the optimal threshold values and ``Caltech 256 image dataset''~\cite{Caltech_image} to evaluate the performance of \textit{Decamouflage} with the selected threshold values in detecting image-scaling attacks. 

We first evaluate the \textit{Decamouflage} detection performance under the white-box setting to validate the feasibility and then under the black-box setting to demonstrate its practicality. The main challenging question we explore in evaluation is as follows:

\begin{center}
    \noindent{\textit{\textbf{RQ. 3: How can we determine an appropriate threshold in\quad\newline white-box or black-box settings?\qquad\qquad\qquad\quad}}}
\end{center}
\vspace{0.08mm}

\begin{figure}[!h]
\centering
\includegraphics[width=0.9\linewidth]{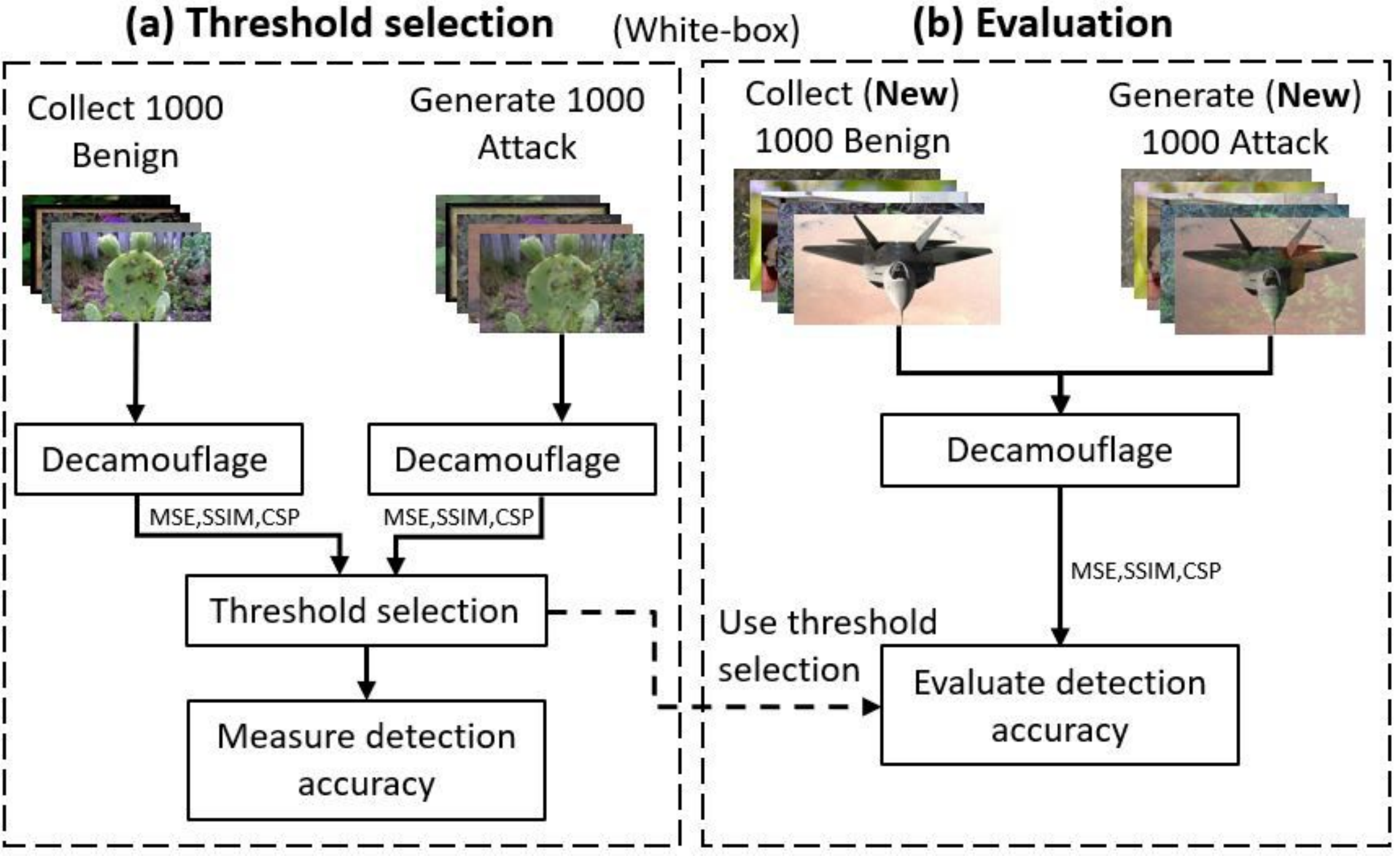}
\caption{White-box setting to validate the feasibility of Decamouflage. (a) Threshold selection, and (b) evaluation.}
\label{fig:White box.}
\end{figure}

\noindent\textbf{White-box setting (Feasibility study):} Following the identified threat model, as presented in Section~\ref{section:TM4.2}, we assume in the white-box setting that we have full access to the attacker's mechanism to mainly demonstrate the feasibility of a detection method. In this setting, we follow the steps shown in Figure~\ref{fig:White box.}. In the first stage, we randomly select 1000 original images and 1000 target images from the ``NeurIPS 2017 Adversarial Attacks and Defences Competition Track'' image dataset~\cite{kurakin2018adversarial} and generate 1000 attack images by combining original images and target images, respectively; and we select the optimal thresholds with those images (we call them \textit{training dataset}). Next, in the second stage, we randomly select 1000 original images and 1000 target images from the ``Caltech 256 image dataset''~\cite{Caltech_image} and evaluate the detection performance of the detection method with those images (we call them \textit{evaluation dataset}).

To select the optimal threshold value for the scaling detection method presented in Section~\ref{sec: Scaling Detection}, we calculate ${MSE}_{(o, S)}$, ${MSE}_{(a, S)}$, ${SSIM}_{(o, S)}$, and ${SSIM}_{(a, S)}$ for all $o \in O$ and for all $a \in A$. Here, our goal is to show that we can select threshold values to distinguish ${MSE}_{(o, S)}$ and ${SSIM}_{(o, S)}$ from ${MSE}_{(a, S)}$ and ${SSIM}_{(a, S)}$, respectively.

Similarly, to select the optimal threshold value for the filtering detection method presented in Section~\ref{sec: Filtering Detection}, we calculate ${MSE}_{(o, F)}$, ${MSE}_{(a, F)}$, ${SSIM}_{(o, F)}$, and ${SSIM}_{(a, F)}$ for all $o \in O$ and for all $a \in A$.

Again, to select the optimal threshold value for the filtering detection method presented in Section~\ref{sec:sd}, we calculate ${CSP}_{o}$ and  ${CSP}_{a}$ for all $o \in O$ and for all $a \in A$. 
In the following sections, we show that there exists a clear recommended threshold value for each method, and the threshold value can be determined in an automated manner with a training dataset only.

\begin{figure}[!ht]
\centering
\includegraphics[width=0.95\linewidth]{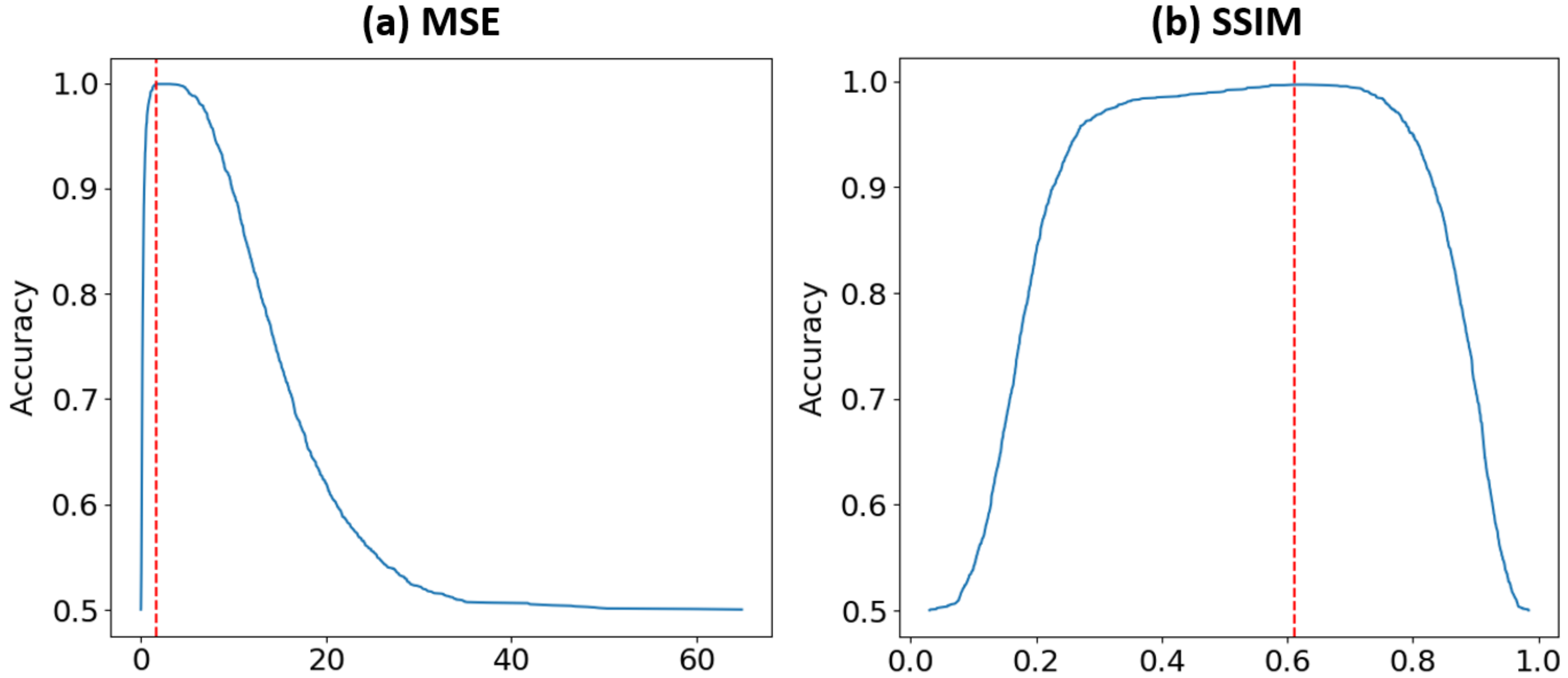}
 \caption{Threshold selection results for the scaling detection method in the white-box setting. The best threshold values are represented by the red dash lines.}
 \label{fig:chooseaccurcy}
\end{figure}

\textbf{Selecting the optimal threshold for a detection method in the white-box setting:} To determine the threshold of a metric $M$ for a detection method in the white-box setting, we developed a gradient descent method that searches for the optimal threshold. The proposed gradient descent method computes the metric values for original images ($M_{original}$) and attack images ($M_{attack}$), respectively, in the training dataset. Next, the gradient descent method picks a metric value from $M_{original}$ and $M_{attack}$, respectively, after ascendingly grading them and determines the threshold as the middle point between them to assess the detection accuracy. This process is repeated until the highest detection accuracy is achieved. As an example, Figure ~\ref{fig:chooseaccurcy} shows the selected threshold result for the scaling detection method. For all detection methods presented in Section~\ref{sec: Potential Detection Methods}, we selected the best thresholds using this gradient descent method.

\begin{figure}[!ht]
\centering
\includegraphics[width=0.8\linewidth]{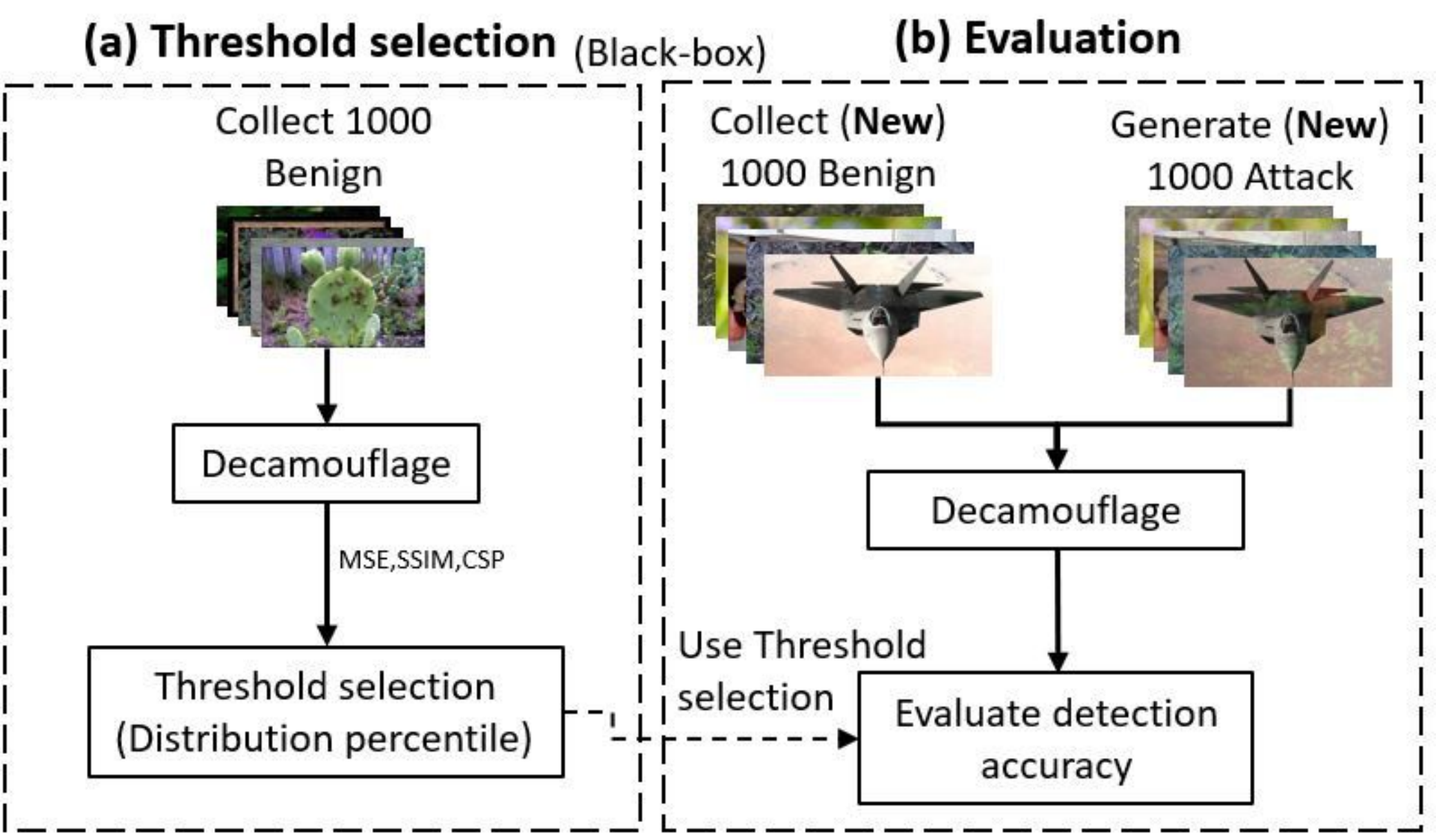}
\caption{Black-box setting to analyze the practicality of Decamouflage. (a) Threshold selection, and (b) evaluation.}
\label{fig:Black box}
\end{figure}

\noindent\textbf{Black-box setting (Practicality study):} The black-box setting evaluates the practicality of a detection method with no assumed knowledge of the attacking mechanism. In this scenario, we need to determine the threshold with benign images alone because there is no access to attack images. The black-box setting also follows two stages shown in Figure~\ref{fig:Black box}. In the first stage, we compute the metric values (i.e., MSE, SSIM, and CSP) with benign images in the training dataset and analyze their statistical distributions to determine the metrics' thresholds. In the second stage, we use the detection methods with the selected thresholds to evaluate the performance of the detection method with the evaluation dataset.

\textbf{Selecting the optimal threshold for the black-box setting:} To determine the threshold of a metric $M$ for a detection method in the black-box setting, we compute the metric values for original images ($M_{original}$) to use the statistical distribution of $M_{original}$, such as its mean and standard deviation. We adopt a percentile of that distribution as a detection boundary and use it as a threshold. \textit{Percentile} is a measure used in statistics indicating the value beyond a given distribution. With the training dataset, we select the optimal percentile of the metrics results 
from their distributions as the threshold achieving the best accuracy results for the detection method.  


The detection accuracy of \textit{Decamouflage} is evaluated with five metrics, accuracy, precision, recall, false acceptance rate (FAR), and false rejection rate (FRR), which are popularly used to evaluate the performance of classifiers.

\begin{itemize}
    \item \textbf{FAR} is the percentage of attack images that are classified as benign images by a detection method.
    \item \textbf{FRR} is the percentage of benign images that are classified as attack images by a detection method. 
    \item \textbf{Accuracy\,(Acc.)} is the percentage of correctly classified images by a detection method.
    \item \textbf{Precision\,(Pre.)} is the percentage of images classified as attack images by a detection method, which are actual attack images.
    \item \textbf{Recall\,(Rec.)} is the percentage of attack images that were accurately classified by a detection method.
\end{itemize} 

In general, while FRR is an indication of detection systems' reliability, FAR shows the security performance. Ideally, both FRR and FAR should be 0\%. Often, a detection system tries to minimize its FAR while maintaining an acceptable FRR as a trade-off, especially under security-critical applications. 


\subsection{Results of the Scaling Detection Method}

\begin{figure}[!h]
\centering
\includegraphics[width=0.95\linewidth]{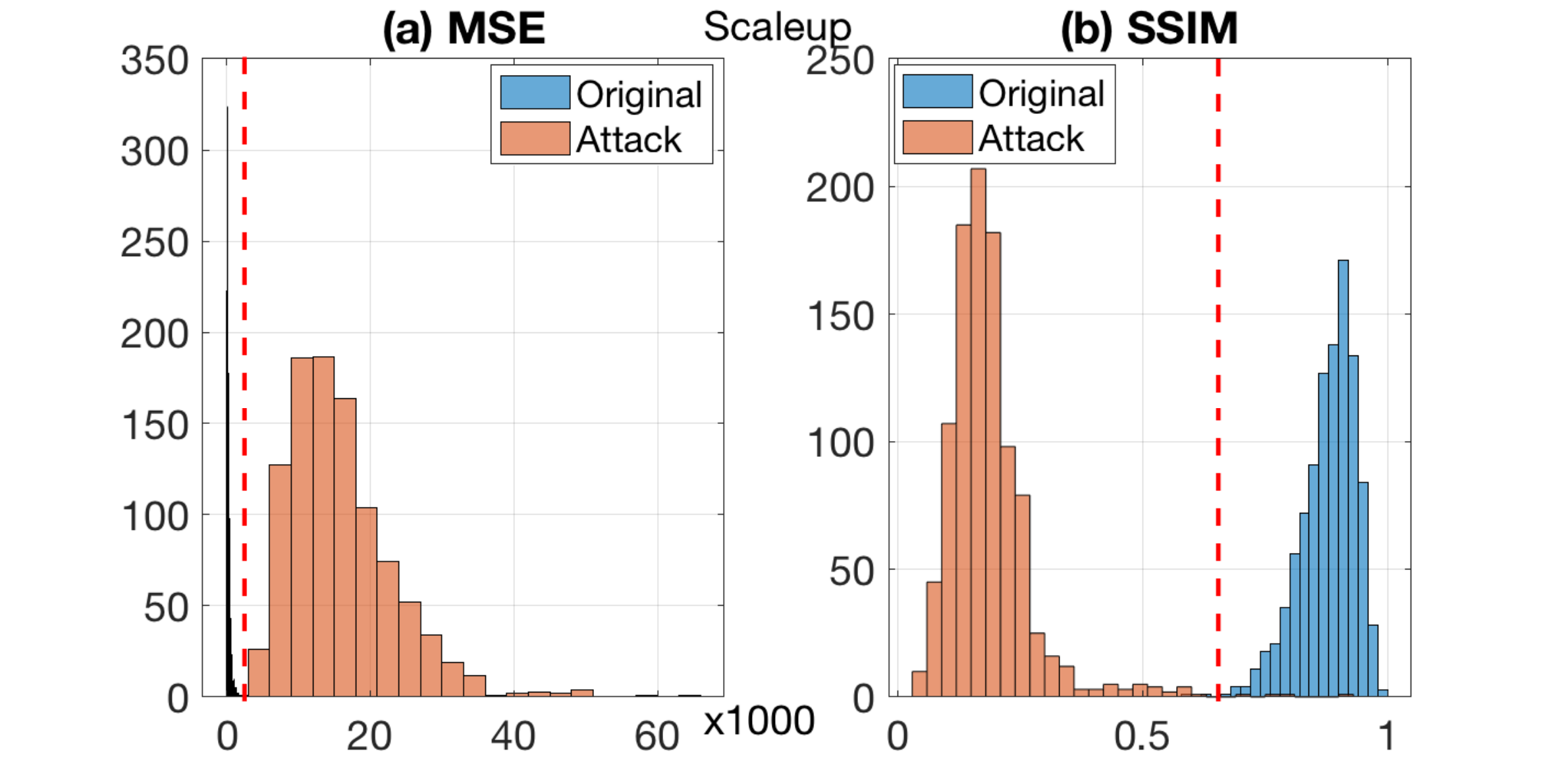}
\caption{Distributions of MSE and SSIM values for the scaling detection method in the white-box setting with 1000 original images and 1000 attack images.}
\label{fig:scaling detection}
\end{figure}

\textbf{Results in the white-box setting:} Figure~\ref{fig:scaling detection} demonstrates that we can find a reasonable threshold (red dashed lines) in both MSE and SSIM to distinguish original images from attack images. We use the gradient descent method to find such thresholds in an automated manner. The selected threshold value for MSE is 1714.96; and the selected threshold value for SSIM is 0.61.

With the selected threshold values, we evaluate the scaling detection method's performance (accuracy, precision, recall, FAR, and FRR) for the evaluation dataset. Table~\ref{tab:scaling wb result} shows that the detection accuracy results of the scaling detection method in the white-box setting. The scaling detection method achieves an accuracy of 99.9\% with FAR of 0.0\% and FRR of 0.1\% for MSE.

\begin{table}[!ht]
\centering
\caption{Results of the scaling detection method in the white-box setting.}
\scalebox{0.88}{
\begin{tabular}{ >{\centering\arraybackslash}m{0.25in} >{\centering\arraybackslash}m{0.3in} >{\centering\arraybackslash}m{0.3in} >{\centering\arraybackslash}m{0.3in} >{\centering\arraybackslash}m{0.3in} >{\centering\arraybackslash}m{0.3in}}\toprule

 & $Acc.$ & $Prec.$ & $Rec.$ & $FAR$ & $FRR$ \\ 
\hline
{$MSE$}  & 99.9\% & 100\% & 99.9\% & 0.0\% & 0.1\% \\

\hline
{$SSIM$} & 99.0\% & 99.7\% & 99.9\% & 0.3\% & 0.1\% \\

\bottomrule
\end{tabular}}
\label{tab:scaling wb result}
\end{table}

\begin{figure}[!h]
\centering
\includegraphics[width=0.95\linewidth]{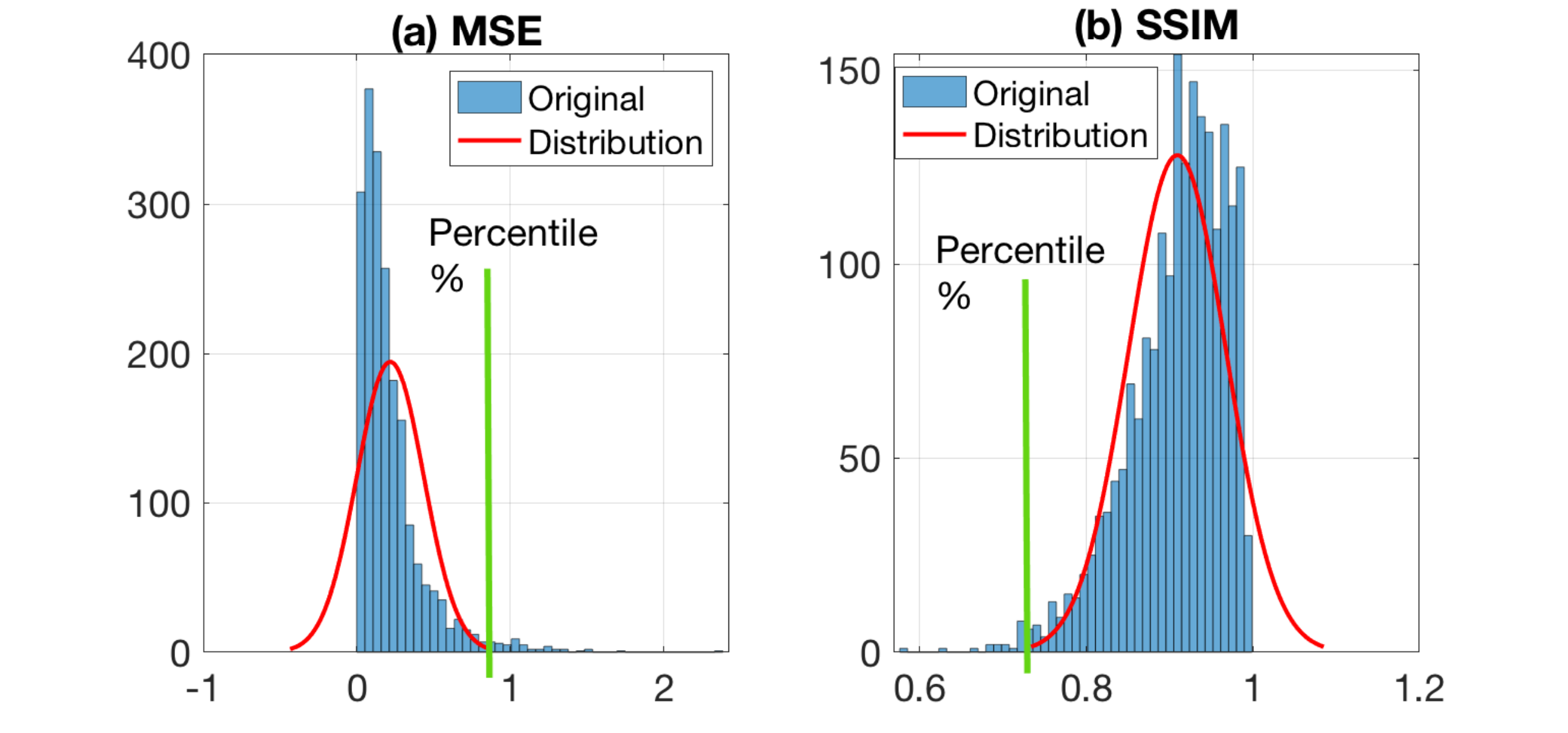}
\caption{Distributions of MSE and SSIM values for the scaling detection method in the black-box setting with 1000 original images. A percentile is represented as a green segment.}
\label{fig:scaling_blackbox}
\end{figure}

\textbf{Results in the black-box setting:} We adopt the \textit{percentile} of the obtained MSE and SSIM distributions built upon 1000 benign images to validate the black-box scenario performance. Figure~\ref{fig:scaling_blackbox} demonstrates that MSE values and the SSIM values follow a normal distribution, respectively, indicating that a percentile-based threshold performs well. As percentile increases, FRR also increases. 



With the three different percentiles (1\%, 2\%, and 3\%), we evaluate the scaling detection method's performance (accuracy, precision, recall, FAR, and FRR) for the evaluation dataset, respectively. Table~\ref{tab:scaling bb result} shows the detection accuracy results of the scaling detection method with the three different percentiles in the black-box setting.  Based on the accuracy results, our recommendation is to use either MSE or SSIM with 1\% percentile. The scaling detection method achieves an accuracy of 99.5\% with FAR of 0.0\% and FRR of 1.0\% for MSE. Similarly, when the percentile is 1\%, the scaling detection method produces the best accuracy of 99.5\% with FAR of 0.0\% and FRR of 1.0\% for SSIM, which are comparable to the results in the white-box setting.

\begin{table}[!ht]
\centering
\caption{Results of the scaling detection method in the black-box setting.}
\scalebox{0.79}{
\begin{tabular}{ >{\centering\arraybackslash}m{0.25in}  >{\centering\arraybackslash}m{0.45in} >{\centering\arraybackslash}m{0.3in} >{\centering\arraybackslash}m{0.3in} >{\centering\arraybackslash}m{0.3in} >{\centering\arraybackslash}m{0.3in} >{\centering\arraybackslash}m{0.3in}
>{\centering\arraybackslash}m{0.4in} >{\centering\arraybackslash}m{0.4in}}\toprule

 & \small \emph{Percentile} & $Acc.$ & $Prec.$ & $Rec.$ & $FAR$ & $FRR$ & $Mean$ & $STD$\\ 
\hline
\multirow{3}{*}{$MSE$}   & 1\%  & 99.5\% & 100.0\% & 99.0\% & 0.0\% & 1.0\% & &\\

                         & 2\%  & 99.0\% & 100.0\% & 98.0\% & 0.0\% & 2.0\% &218.6 & 217.6\\

                         & 3\%  & 98.5\% & 100.0\% & 97.1\% & 0.0\% & 3.0\% & & \\
\hline
\multirow{3}{*}{$SSIM$}  & 1\%  & 99.5\% & 100.0\% & 99.0\% & 0.0\% & 1.0\% & &\\

                         & 2\%  & 99.0\% & 100.0\% & 98.0\% & 0.0\% & 2.0\% & 0.91 & 0.59\\

                         & 3\%  & 98.5\% & 100.0\% & 97.0\% & 0.0\% & 3.0\% & &\\

\bottomrule
\end{tabular}}
\label{tab:scaling bb result}
\end{table}

\subsection{Results of the Filtering Detection Method}
\begin{figure}[!h]
\centering
\includegraphics[width=1\linewidth]{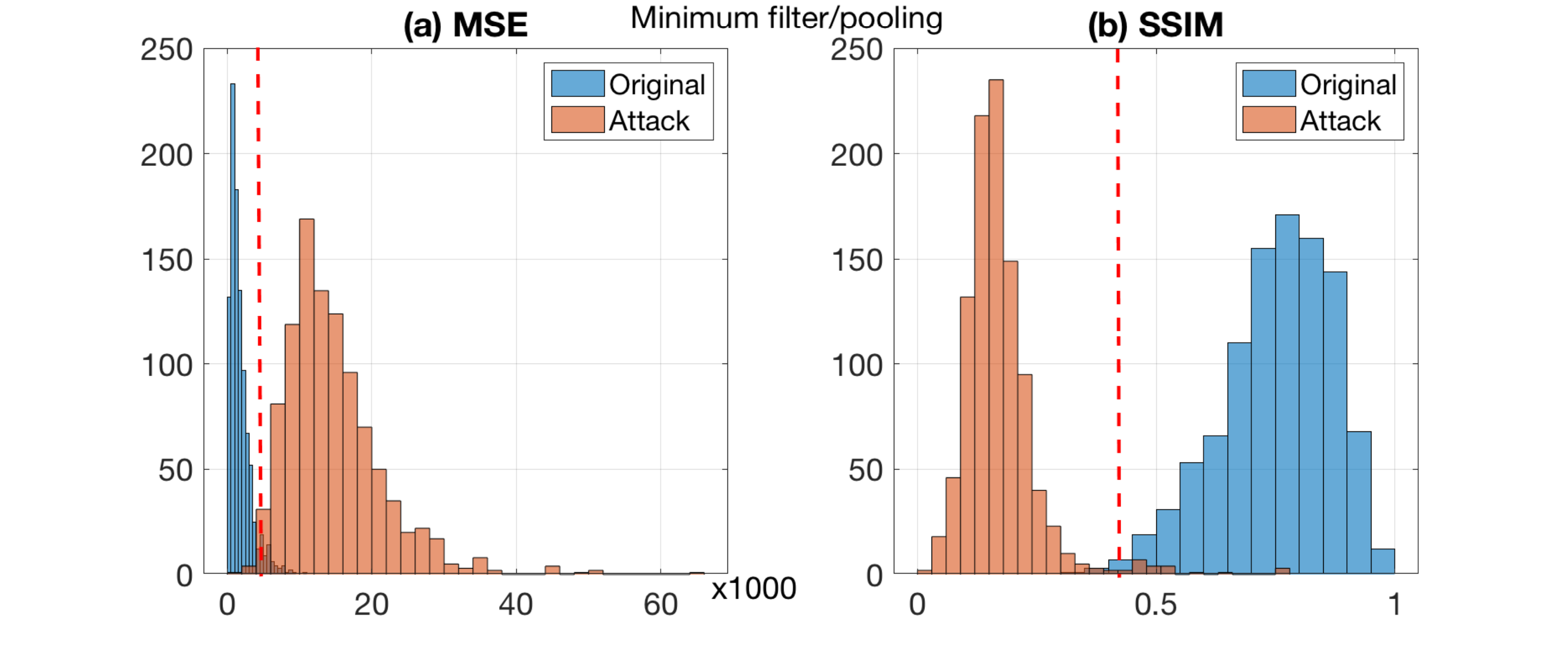}
\caption{Distributions of MSE and SSIM values for the filtering detection method in the white-box setting with 1000 original images and 1000 attack images.} 
\label{fig:max/min_whitebox}
\end{figure}

\textbf{Results in the white-box setting:} Figure~\ref{fig:max/min_whitebox} demonstrates that we can find a reasonable threshold (red dashed lines) in both MSE and SSIM to distinguish original images from attack images even though there exist some overlapped part between them in MSE. Again, we use the gradient descent method to find such thresholds in an automated manner. The selected threshold value for MSE is 5682.79; and the selected threshold value for SSIM is 0.38.

With the selected threshold values, we evaluate the filtering detection method's performance (accuracy, precision, recall, FAR, and FRR) for the evaluation dataset. Table~\ref{tab:filtering wb result} shows that the detection accuracy results of the filtering detection method in the white-box setting. The filtering detection method achieves an accuracy of 99.3\% with FAR of 1.3\% and FRR of 0.2\% for SSIM.

\begin{table}[!ht]
\centering
\caption{Results of the filtering detection method in the white-box setting.}
\scalebox{0.9}{
\begin{tabular}{ >{\centering\arraybackslash}m{0.25in}  >{\centering\arraybackslash}m{0.3in} >{\centering\arraybackslash}m{0.3in} >{\centering\arraybackslash}m{0.3in} >{\centering\arraybackslash}m{0.3in} >{\centering\arraybackslash}m{0.3in} >{\centering\arraybackslash}m{0.3in}}\toprule

& $Acc.$ & $Prec.$ & $Rec.$ & $FAR$ & $FRR$\\ 
\hline
{$MSE$} & 98.6\%  &97.5\%  &99.2\%  &2.5\%  &0.8\% \\ 

\hline

{$SSIM$} & 99.3\%  &98.7\%  &99.7\%  &1.3\%  &0.2\% \\

\bottomrule
\end{tabular}}

\label{tab:filtering wb result}
\end{table}

\begin{figure}[!h]
\centering
\includegraphics[width=1\linewidth]{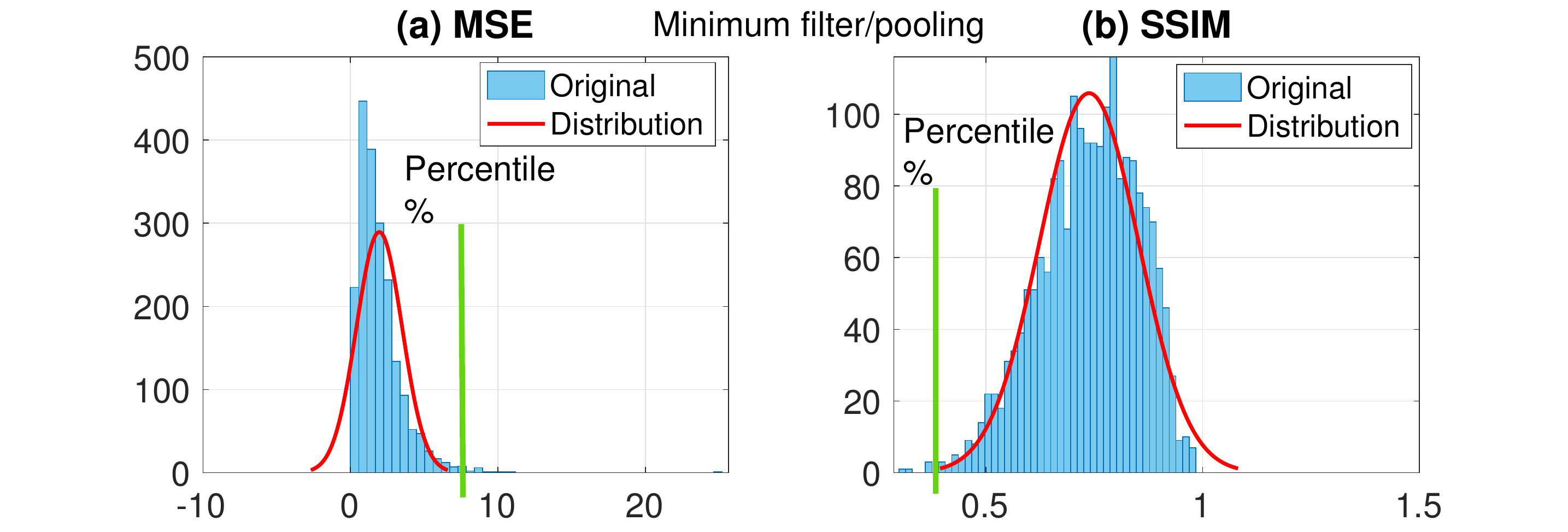}
\caption{Distributions of MSE and SSIM values for the filtering detection method in the black-box setting with 1000 original images. A percentile is represented as a green segment.}
\label{fig:max/min_blackbox}
\end{figure}

\textbf{Results in the black-box setting:} We adopt the \textit{percentile} of the obtained MSE and SSIM distributions built upon 1000 benign images to validate the black-box scenario performance. Figure~\ref{fig:max/min_blackbox} demonstrates that MSE values and the SSIM values follow a normal distribution, respectively, indicating that a percentile-based threshold performs well.   

With the three different percentiles (1\%, 2\%, and 3\%), we evaluate the filtering detection method's performance (accuracy, precision, recall, FAR, and FRR) for the evaluation dataset, respectively. Table~\ref{tab:filtering Min bb result} shows the detection accuracy results of the filtering detection method with the three different percentiles in the black-box setting. Based on the accuracy results, our recommendation is to use SSIM with 1\% percentile. In this case, the filtering detection method achieves an accuracy of 99.2\% with FAR of 0.6\% and FRR of 1.0\% for SSIM.

\begin{table}[!h]
\centering
\caption{Results of the filtering detection method in black-box setting.}
\scalebox{0.79}{
\begin{tabular}{ >{\centering\arraybackslash}m{0.25in}  >{\centering\arraybackslash}m{0.45in} >{\centering\arraybackslash}m{0.3in} >{\centering\arraybackslash}m{0.3in} >{\centering\arraybackslash}m{0.3in} >{\centering\arraybackslash}m{0.3in} >{\centering\arraybackslash}m{0.3in}
>{\centering\arraybackslash}m{0.4in} >{\centering\arraybackslash}m{0.4in}}\toprule

 & \small \emph{Percentile} & $Acc.$ & $Prec.$ & $Rec.$ & $FAR$ & $FRR$ & $Mean$ & $STD$\\ 
\hline
\multirow{3}{*}{$MSE$}   & 1\%  &98.4\%  &97.8\%  &98.9\%  &2.2\%  &1.0\%  & &\\

                         & 2\%  &98.5\%  &99.0\%  &98.1\%  &0.9\%  &2.0\%  &1952.32  &1543.27 \\

                         & 3\%  &98.2\%  &99.4\%  &97.1\%  &0.5\%  &3.0\%  & & \\
\hline
\multirow{3}{*}{$SSIM$}  & 1\%  &99.2\%  &99.3\%  &98.9\%  &0.6\%  &1.0\%  & &\\

                         & 2\%  &98.7\%  &99.4\%  &98.0\%  &0.5\%  &2.0\%  &0.74  &0.11 \\

                         & 3\%  &98.2\%  &99.4\%  &96.9\%  &0.5\%  &3.0\%  & & \\

\bottomrule
\end{tabular}}
\label{tab:filtering Min bb result}
\end{table}

\subsection{Results of the Steganalysis Detection Method}
\begin{figure}[!h]
\centering
\includegraphics[width=0.66\linewidth]{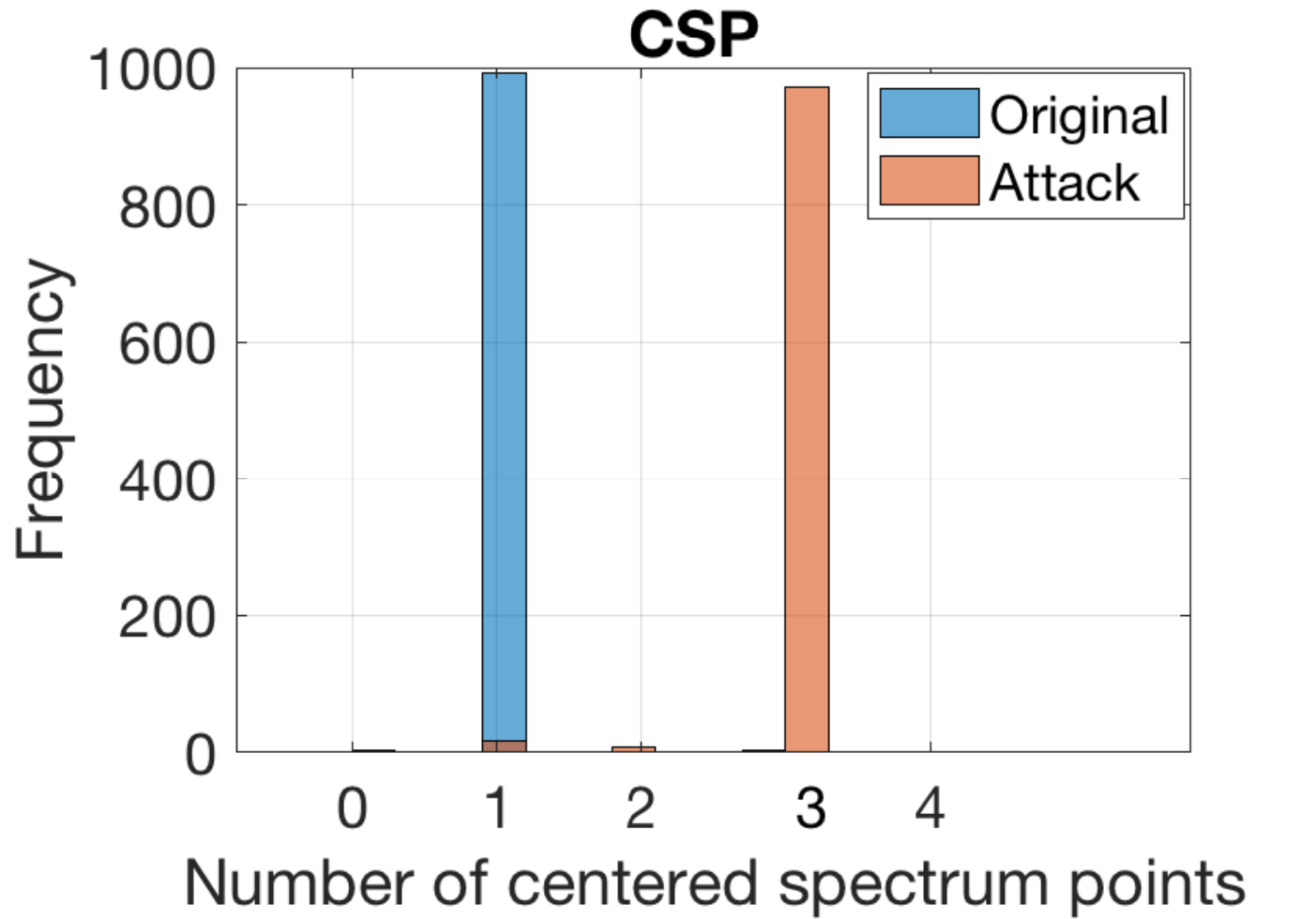}
\caption{Distributions of CSP values for the steganalysis detection method in the white-box setting with 1000 original images and 1000 attack images.}
\label{fig:stego_white_box}
\end{figure}

\textbf{Results in the white-box setting:} Figure~\ref{fig:stego_white_box} shows that 99.3\% of original images have 1 CSP, whereas 98.2\% of attack images have more than 1 CSP, indicating that we can clearly distinguish them if we set the CSP threshold to 2. 

With the CSP threshold of 2, we evaluate the steganalysis detection method's performance (accuracy, precision, recall, FAR, and FRR) for the evaluation dataset. Table~\ref{tab:Frequency white box} shows that the detection accuracy results of the steganalysis detection method in the white-box setting. The steganalysis detection method achieves an accuracy of 98.9\% with FAR of 0.3\% and FRR of 1.7\%.

\begin{table}[!ht]
\centering
\caption{Results of the steganalysis detection method in the white-box setting.}
\scalebox{0.88}{
\begin{tabular}{ >{\centering\arraybackslash}m{0.3in}  >{\centering\arraybackslash}m{0.3in} >{\centering\arraybackslash}m{0.3in} >{\centering\arraybackslash}m{0.3in} >{\centering\arraybackslash}m{0.3in}
>{\centering\arraybackslash}m{0.3in}}\toprule

& $Acc.$ & $Prec.$ & $Rec.$ & $FAR$ & $FRR$ \\ 
\hline
$CSP$ & 98.9\%  &99.7\%  &98.2\%  &0.3\% &1.7\% \\

\bottomrule
\end{tabular}}
\label{tab:Frequency white box}
\end{table}

\textbf{Results in the black-box setting:} Interestingly, we do not need to analyze the CSP distribution of original images in the steganalysis detection method, unlike the other detection methods. Based on our observation of the white-box setting experiments, we surmise that the attack images generated by image-scaling attacks inherently have multiple centered spectrum points. Therefore, we use a fixed threshold of 2 for CSP in the steganalysis detection method regardless of original and attack images. Consequently, we can reduce the cost of determining thresholds in the steganalysis detection method. If we use 2 for the CSP threshold, the steganalysis detection method achieves an accuracy of 98.9\% with FAR of 0.3\% and FRR of 1.7\%, which are the same as the results in the white-box setting.

\subsection{Run-time Overhead and Ensemble Approach}

\textbf{Run-time overhead:} As the threshold determination is performed offline, we focus on the most concerning overhead --- run-time overhead in a real-time situation. In other words, how long the plug-in \textit{Decamouflage} system takes from getting an input image until producing the detection decision. We implement \textit{Decamouflage} in Python 3. We use a PC with an Intel Core i5-7500 CPU (3.41GHz) and 8GB memory in all our experiments. Table~\ref{tab:Time_overhead} details the run-time overhead of \textit{Decamouflage} system. The decision requires between $3$ and $174$ millisecond/image on average.

Furthermore, each method's standard deviation is small, indicating that it takes a similar time regardless of images. Those measurement results demonstrate that \textit{Decamouflage} can be deployed for real-time detection. Notably, the steganalysis detection method can be deployed to detect image-scaling attacks efficiently without the threshold setup process.

\begin{table}[!ht]
\renewcommand{\arraystretch}{1.3}
\centering
\caption{Run-time overheads of detection methods}
\scalebox{0.8}{
\begin{tabular}
{ >{\centering\arraybackslash}m{1.1in} || >{\centering\arraybackslash}m{0.5in}  | >{\centering\arraybackslash}m{0.7in} |>{\centering\arraybackslash}m{0.7in}}
\hlineB{2.5}
Method & Matrix & Run-time overhead (millisecond) & Standard deviation (millisecond)\\ 
\hline
\multirow{2}{*}{$Scaling$} & MSE & 11 & 5\\
        & SSIM & 137 & 4\\
\hline
\multirow{2}{*}{$Filtering$} & MSE & 11 & 3\\
                                & SSIM & 174 & 6\\
\hline
$Steganalysis$ & CSP & 3 & 1\\
\hlineB{2.5}
\end{tabular}}
\label{tab:Time_overhead}
\end{table}



\textbf{Ensemble approach:} We showed that each of the three detection methods in Section~\ref{sec: Potential Detection Methods} produced a high detection accuracy against image-scaling attacks. In this paragraph, we discuss the possibility of an ensemble approach of those methods to improve the reliability and detection accuracy. We can develop a simple ensemble model based on a majority voting rule of multiple detection methods. Its advantages are that (1) it achieves better and stable results, and (2) it hardens adaptive attacks that could be effective against a particular detection method. Table \ref{tab:Ensemble} shows the detailed experimental results, where the performance of both the white-box and black-box ensemble models are evaluated. 

\begin{table}[!ht]
\centering
\caption{Result of Decamouflage system as an ensemble model. The black-box and white-box settings both  demonstrate promising results.}
\scalebox{0.9}{
\begin{tabular}{ >{\centering\arraybackslash}m{1.2in} >{\centering\arraybackslash}m{0.3in} >{\centering\arraybackslash}m{0.3in} >{\centering\arraybackslash}m{0.3in} >{\centering\arraybackslash}m{0.3in} >{\centering\arraybackslash}m{0.3in}
}\toprule

 & $Acc.$ & $Prec.$ & $Rec.$ & $FAR$ & $FRR$\\ 
\hline
{White-box ensemble}   &99.9\% &99.8\%  &100.0\%  &0.2\% &0.0\%\\
\hline
{Black-box ensemble}   &99.8\% &99.8\% &99.9\%  &0.2\% &0.1\%\\

\bottomrule
\end{tabular}}
\label{tab:Ensemble}
\end{table}

In the white-box setting, \textit{Decamouflage} achieves an accuracy of 99.9\% with FAR of 0.2\% and FRR of 0.0\%, indicating that it does not classify any original images mistakenly into attack images with a minimal false acceptance. Moreover, even in the black-box setting, \textit{Decamouflage} can produce highly accurate outputs achieving an accuracy of 99.8\% with FAR of 0.2\% and FRR of 0.1\%, which slightly outperforms the best configuration of each detection method.\\






\noindent{\bf Summary:} As an answer to RQ. 3, we present how to determine an appropriate threshold in the white-box and black-box settings. In the white-box setting, we specially develop a gradient descent method that searches for each metric's optimal threshold across the dataset of benign and attack images and uses that threshold against an unseen dataset. In the black-box setting, we adopt the percentile as a detection boundary after analyzing the statistical distribution of original images in a metric.


%% file: Section/6_Discussion.tex
\section{Discussions}
\label{sec:EM}

\vspace{0.2cm}
\noindent{\bf Considerations for adaptive attacks:} \textit{Decamouflage} is built upon the three detection methods: scaling, filtering, and steganalysis. In fact, our experimental results demonstrate that {\it each} of the three methods is sufficiently accurate to detect image-scaling attacks and thus can be individually opted for deployment. However, those detection methods can be incorporated together to work in an ensemble manner to harden the adaptive attacks: an attacker now has to bypass them concurrently. Quiring {\it et al.}~\cite{quiring2020backdooring} demonstrated that by developing an adaptive attack to Xiao {\it et al.}'s~\cite{xiao2019seeing} initial mitigation strategy of using an image histogram. Considering this kind of possibility of adaptive attacks, \textit{Decamouflage} has been developed for defense-in-depth of the image-scaling attack detection system.

\vspace{0.2cm}
\noindent{\bf Robustness of image similarity metrics:}
To quantify the difference between the input image and its rescaled or filtered counterpart, we suggested two metrics: MSE and SSIM (see Section~\ref{sec:residual}). We believe that MSE-based detection methods' performance could deteriorate with highly distorted images because MSE relies on measuring the absolute errors, whereas SSIM-based detection can take luminance, contrast, and structure of images into consideration~\cite{silva2007quantifying}. After all, it would be more robust against such distorted images. Interestingly, unlike MSE and SSIM, we observed that PSNR could be ineffective in showing a threshold to distinguish benign images from attack images even though PSNR is also popularly used to calculate the physical difference between the two images (see Appendix~\ref{sec: PSNR ineffective as a threshold}). We surmise that this is due to peak errors that can significantly affect PSNR values. On the other hand, MSE relies on the cumulative squared errors that soften the difference between the benign and its rescaled or filtered counterpart into {\it lower level}, which can reduce the effects of peak errors.  

\vspace{0.2cm}
\noindent{\bf Characteristics of the attack images that cannot be detected by Decamouflage:} We analyze the attack images that are falsely accepted as benign images by \textit{Decamouflage}. Table~\ref{tab:misclassified} and Appendix \ref{append:misclassified_sample} show a few representative examples of such attack images. Therefore, attackers can try to generate such attack images using adversarial machine learning techniques for bypassing \textit{Decamouflage} intentionally. However, we found that it would be very challenging to generate attack images that cannot be detected by \textit{Decamouflage} and are still effective. We analyzed the attack images that \textit{Decamouflage} failed to detect with commercial cloud-based computer vision services that deploy the state-of-the-art machine learning models including Microsoft Azure\footnote{\url{https://azure.microsoft.com/en-us/services/cognitive-services/computer-vision/?v=18.05}}, Baidu\footnote{\url{https://ai.baidu.com/tech/imagerecognition/fine_grained}}, and Tencent\footnote{\url{https://ai.qq.com/product/visionimgidy.shtml}}. We observed that most of such attack images were not recognized as attackers' target images. For example, as presented in Table~\ref{tab:misclassified}, both attack images were not classified as the target images by all the tested three computer vision services ---losing their attacking purpose.

\begin{table*}[!h]
\renewcommand{\arraystretch}{1.3}
\centering
\caption{Example attack images that are mistakenly accepted by \textit{Decamouflage}. Those images have been misclassified as different objects by three computer vision classifiers (Azure, Baidu, and Tencent), indicating that while those attack images may pass the system, they might lose the attacker's original purpose.}

\scalebox{0.76}{
\begin{tabular}{|m{0.10\linewidth}|m{0.19\linewidth}|m{0.19\linewidth}|m{0.19\linewidth}|m{0.19\linewidth}|}
\hline
&\multicolumn{2}{m{0.38\linewidth}|}{\textbf{\textit{\qquad\qquad\qquad Original\,vs\,Attack}}}&\multicolumn{2}{m{0.38\linewidth}|}{\textbf{\textit{\qquad\qquad\qquad Original\,vs\,Attack}}}\\
\hline
\textbf{\textit{\quad\, Image}}&\includegraphics[height=0.5in,width=1in, margin=11pt 1.5pt 10pt 1.5pt,valign=m]{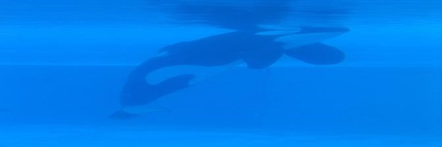}&\includegraphics[height=0.5in,width=1in, margin=11pt 1.5pt 10pt 1.5pt,valign=m]{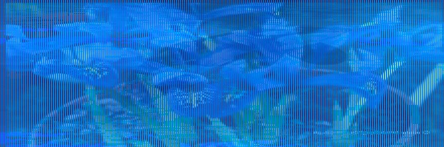}&\includegraphics[height=0.5in,width=1in, margin=11pt 1.5pt 10pt 1.5pt,valign=m]{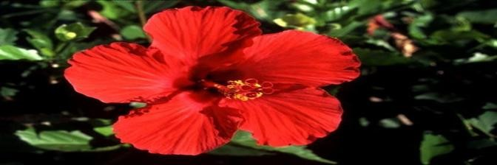}&\includegraphics[height=0.5in,width=1in, margin=11pt 1.5pt 10pt 1.5pt,valign=m]{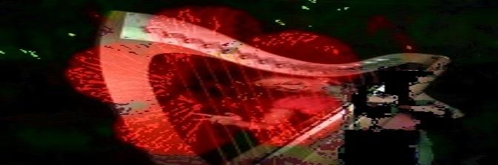}\\
\hline
\textbf{\textit{\quad\, Azure}}&\qquad\quad\quad 42.3\%&\qquad\quad\quad 47.2\%&\qquad\quad\quad 99.4\%&\,68.6\%\,text,\,66.1\%\,glass,\\
&\,\,\,A\,fish\,swim\,under\,water&\,\,\,\, A\,blue\,background & \qquad\quad\, A\,flower &\qquad\, 61.8\%\,soft\,drink\\
\hline
\textbf{\textit{\quad\, Baidu}}&\qquad\quad\quad 99.8\%&\qquad\quad\quad 99.3\%&\qquad\quad\quad 62.2\%&\quad Subject\,not\,detected\\
&\,\,\,\,\qquad Killer\,whale&\,\,\,\,\qquad Non-animal&\qquad\quad\, Hibiscus &\\
\hline
\textbf{\textit{\quad Tencent}}&\,\,\,25\%\,animal,\,53\%\,water,&18\% \,night,18\%\,screenshot&\,\,65\%\,flower,\,25\%\,branches&\,12\%\,night,\,14\%\,cave,\,16\%\\
&\qquad\quad\; 35\%\,fish& &\qquad\quad\, and\,leaves &\,rock,\,15\%\,water,\,14\%\,light\\
\hline
\end{tabular}}
\label{tab:misclassified}
\end{table*}

%% file: Section/7_Related_work.tex
\section{Related Work}
Several techniques have been proposed in the literature to violate the security of neural network models, as detailed in \cite{biggio2018wild,papernot2018sok}. In recent years, many new attack and defense techniques~\cite{biggio2013evasion,carlini2017towards,li2018textbugger,lecuyer2019certified,quiring2019misleading} have been developed in the area of adversarial machine learning field. Unlike the image-scaling attack introduced by Xiao {\it et al.}~\cite{xiao2019seeing}, adversarial examples are neural network dependent. In the white-box setting, they are specifically designed based on the knowledge about the model parameters such as weights and inputs to trick a model into making an erroneous prediction. In the black-box setting, the adversary still needs to look at the model output in many iterations to generate an adversarial sample. In contrast, the image-scaling attack is agnostic to feature extraction and learning models because it targets the early preprocessing pipeline --- rescaling operation. The image-scaling attack also greatly facilitates data poisoning attacks to insert a backdoor into the CNN model~\cite{gao2020backdoor}. Quiring {\it et al.}~\cite{quiring2020backdooring} explored this possibility explicitly. 



As far as we know, defense mechanisms against image scaling attacks were only investigated by Quiring \mbox{{\it et al.} \cite{quiring2020adversarial}}. They suggested two prevention mechanisms to prohibit the scaling function from injecting the desired attack image. However, their suggested techniques have a few limitations, as mentioned in Section~\ref{sec:intro}, such as incompatibility with existing scaling algorithms and side-effects of degrading the input image quality using the image reconstruction method. In this paper, we propose a novel image-scaling attack detection framework called \emph{Decamouflage} to overcome these limitations. 



%% file: Section/8_Conclusion.tex
\section{Conclusion}
We present \textit{Decamouflage} to detect image-scaling attacks, which can affect many computer vision applications using image-scaling functions. We explored the three promising detection methods: scaling, filtering, and steganalysis, which can be individually deployed or incorporated together as an ensemble solution. We performed extensive evaluations with two independent datasets, demonstrating the effectiveness of \textit{Decamouflage} (see more examples in Appendix~\ref{append:Scaling detection method Visual samples}, \ref{append:Filtering detection method visual samples}, and \ref{append:Steganalysis detection method visual samples}). For each detection method of \textit{Decamouflage}, we suggest the best metric and thresholds maximizing the detection accuracy. In particular, the steganalysis detection method can be efficiently used with a fixed threshold for CSP regardless of datasets. Our detection solutions can be robust and effective as an ensemble solution with those detection methods. In the white-box setting (for the feasibility study), \textit{Decamouflage} achieves an accuracy of 99.9\% with FAR of 0.2\%, and FRR of 0.0\%. Even in the black-box setting (for the practicality study), \textit{Decamouflage} achieves an accuracy of 99.8\% with FAR of 0.2\%, and FAR of 0.1\%. Moreover, the run-time overhead evaluation shows that the \textit{Decamouflage} is also acceptable to be deployed for real-time online detection.  


%% file: Section/9_Appendix.tex
\appendix
\section{Possibility of PSNR as a metric for Decamouflage}
\label{sec: PSNR ineffective as a threshold}

PSNR computes the ratio between the maximum possible power of an image and the power of corrupting noise that affects the quality of its representation. The PSNR can be defined as follows:
\begin{equation}\label{eq:psnr}
    PSNR = 10\log_{10}\left (\frac{(L-1)^2}{MSE}\right)
\end{equation}  
where $L$ is the number of maximum possible intensity levels (pixel values) of an image which then divided by the mean square root.

\begin{figure}[!h]
\centering
\includegraphics[width=0.5\linewidth]{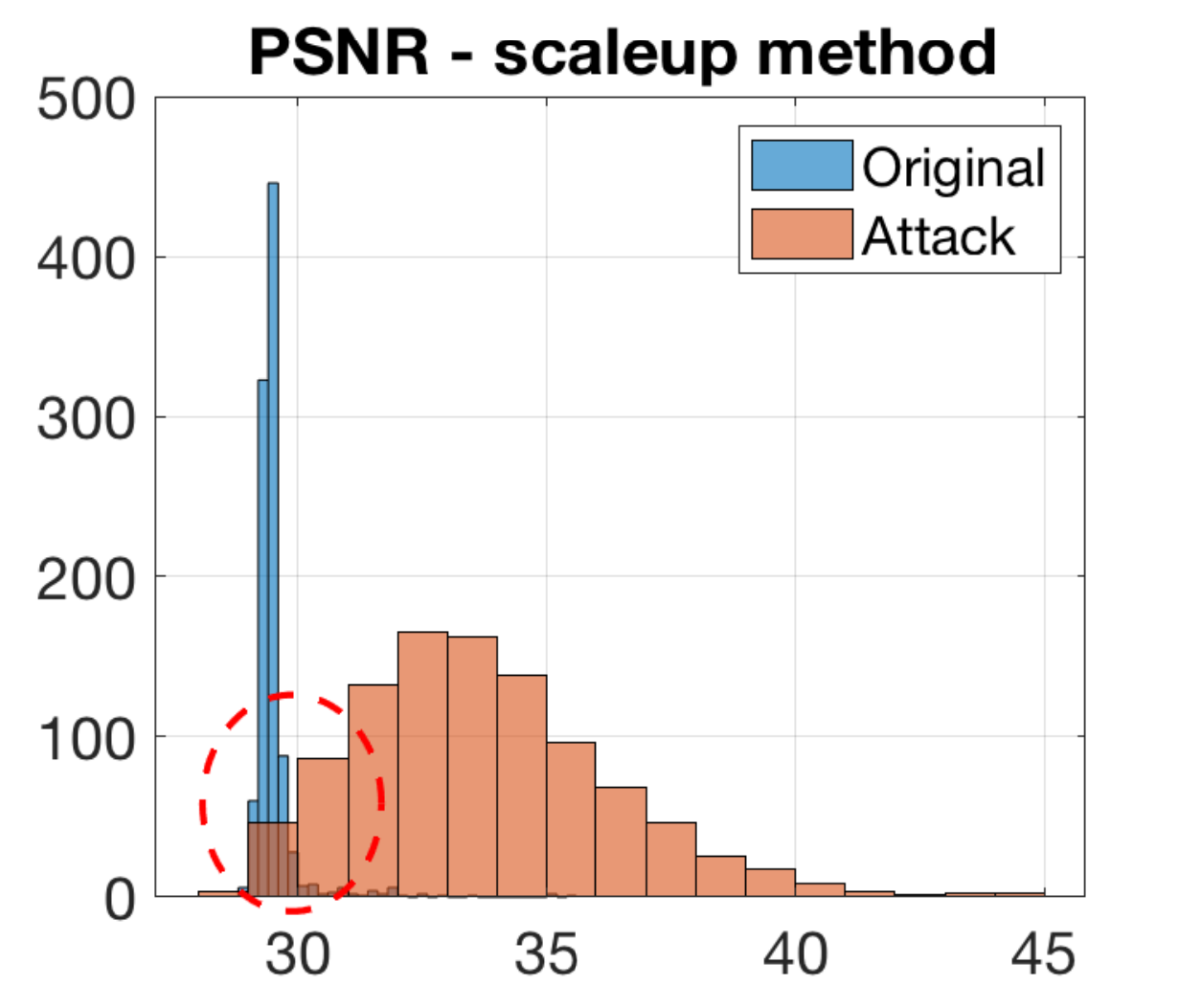}
\caption{Histogram results of PSNR obtained from 1000 benign and 1000 attack images for the scaling detection method in the white-box setting. The PSNR values obtained from 1000 benign images are highly overlapped with the 1000 attack images.} 
\label{fig:scaling_psnr}
\end{figure}

We found that PSNR would not be recommendable in the scaling detection method presented in Section~\ref{sec: Scaling Detection}. Figure~\ref{fig:scaling_psnr} shows that the PSNR values obtained from 1000 benign images are highly overlapped with the 1000 attack images. Therefore, we do not recommend using PSNR for the scaling detection method.

\begin{figure}[!h]
\centering
\includegraphics[width=0.5\linewidth]{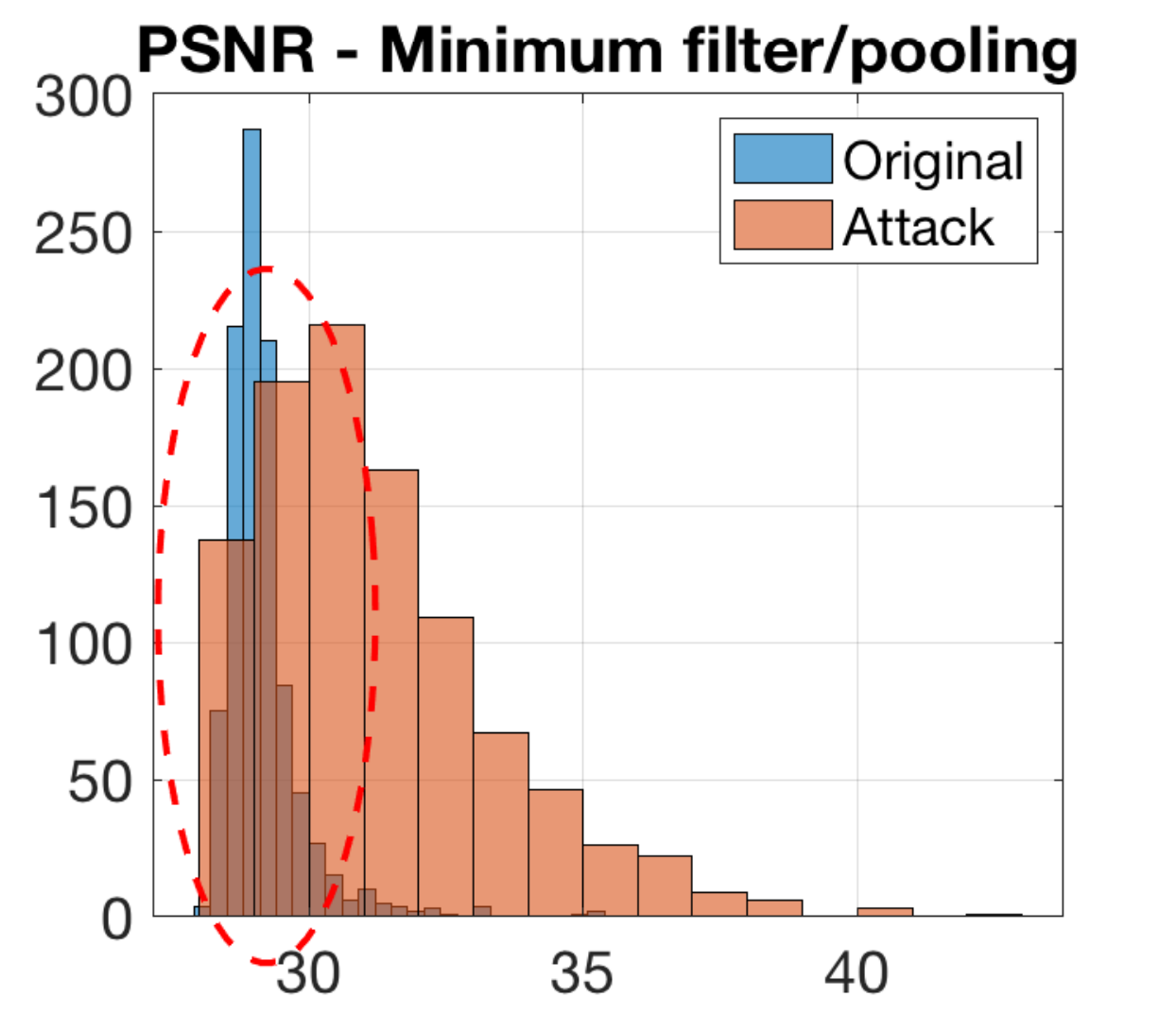}
\caption{Histogram results of PSNR obtained from 1000 benign and 1000 attack images for the filtering detection method in the white-box setting. The PSNR values obtained from 1000 benign images are highly overlapped with the 1000 attack images in minimum filter.} 
\label{fig:filter_psnr}
\end{figure}

Similarly, Figure~\ref{fig:filter_psnr} demonstrates that PSNR is not recommendable for the filtering detection method presented in Section~\ref{sec: Filtering Detection}.

\section{More examples from the ones that got away}
\label{append:misclassified_sample}

\begin{figure}[!ht]
\centering
\includegraphics[width=0.9\linewidth]{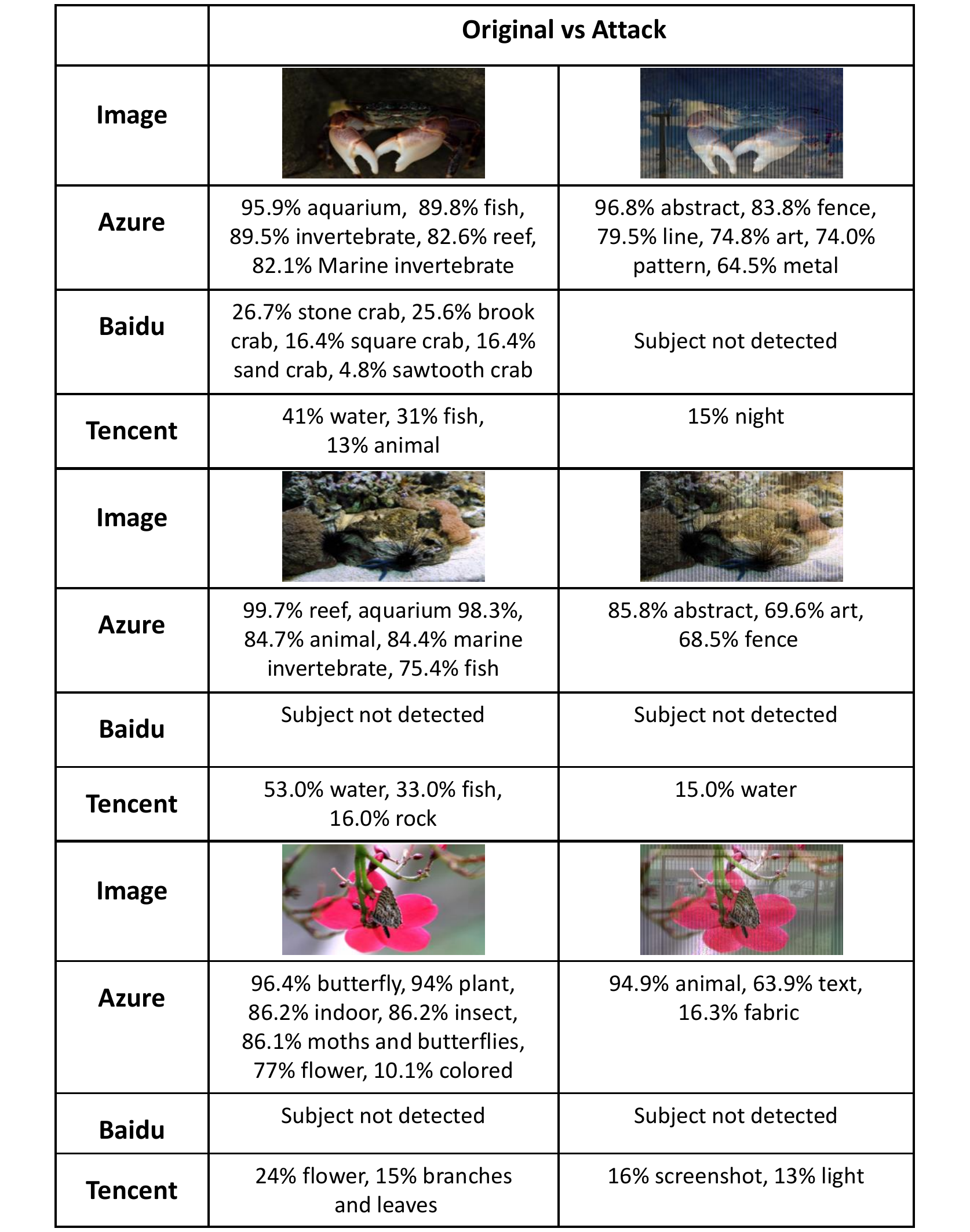}
\caption{More attack image examples that are mistakenly accepted by \textit{Decamouflage}. They have been classified as different objects by three computer vision classifiers (Azure, Baidu, and Tencent). They also indicate that while those attack images may pass the system, but they might also lose the attack purpose.}
\label{fig:misclassifiedd}
\end{figure}

Figure \ref{fig:misclassifiedd} provides more examples as misclassified as benign by Decamouflage system. The results also suggest that while they have been misclassified by our system, attack image results in being misclassified not to its targeted label by various computer vision classifiers which means the attack loses the purpose as well.

\vspace{-1mm}

\vspace{-3mm}
\section{Scaling detection method Visual samples}
\label{append:Scaling detection method Visual samples}

\begin{figure*}[!ht]
\centering
\includegraphics[width=0.8\linewidth]{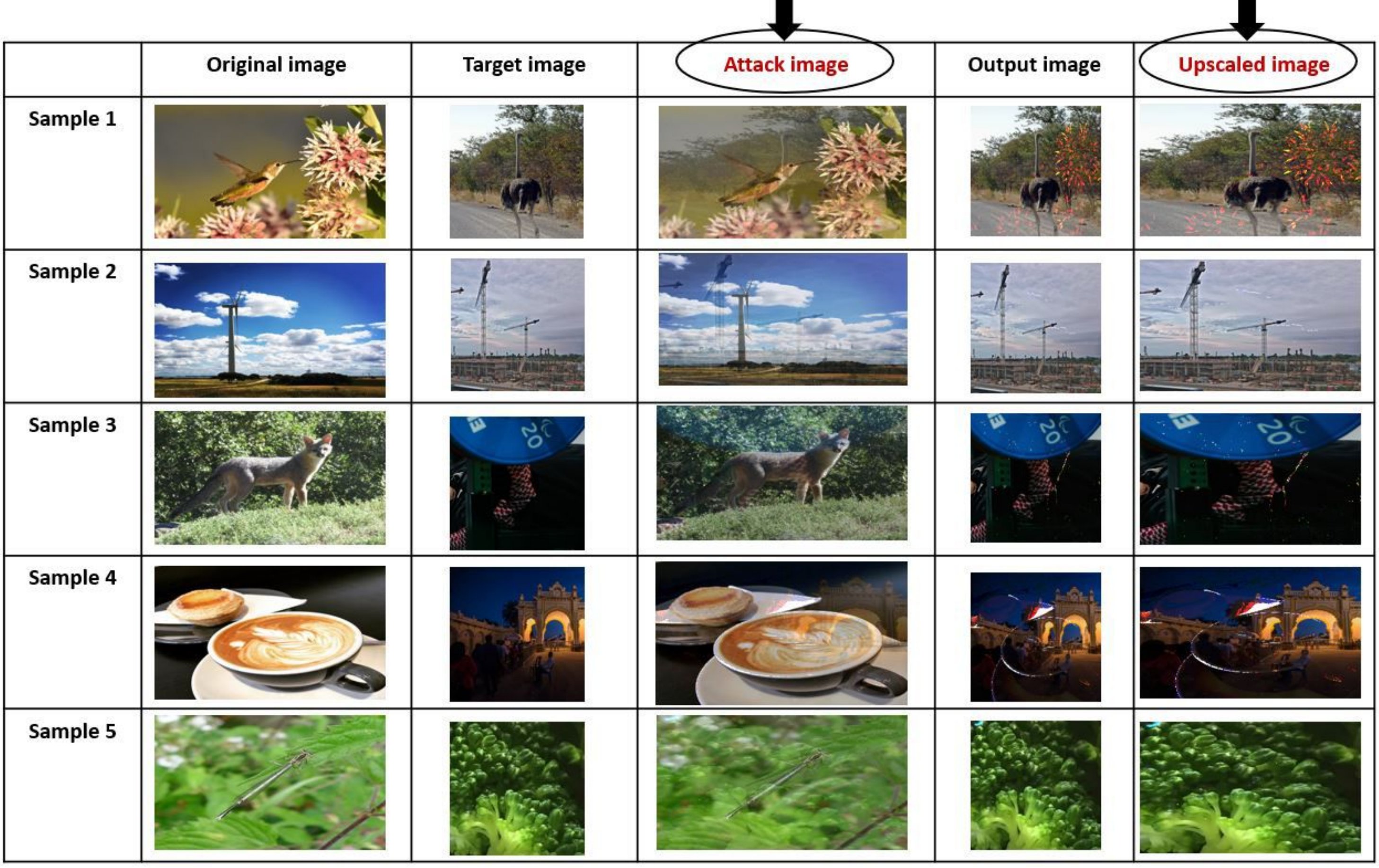}
\caption{More examples from our scaling detection method. They consistently show notable differences between the \textit{attack images} and \textit{upscaled images}. This difference is conveniently quantified by various metrics such as MSE and SSIM.}
\label{fig:scaling_images}
\end{figure*}

Figure \ref{fig:scaling_images} presents additional visual examples that demonstrate the scaling detection method. Our \textit{Decamouflage} system is able to quantify the difference using both MSE and SSIM metrics.

\section{Filtering detection method visual samples}
\label{append:Filtering detection method visual samples}
\begin{figure*}[!ht]
\centering
\includegraphics[width=0.9\linewidth]{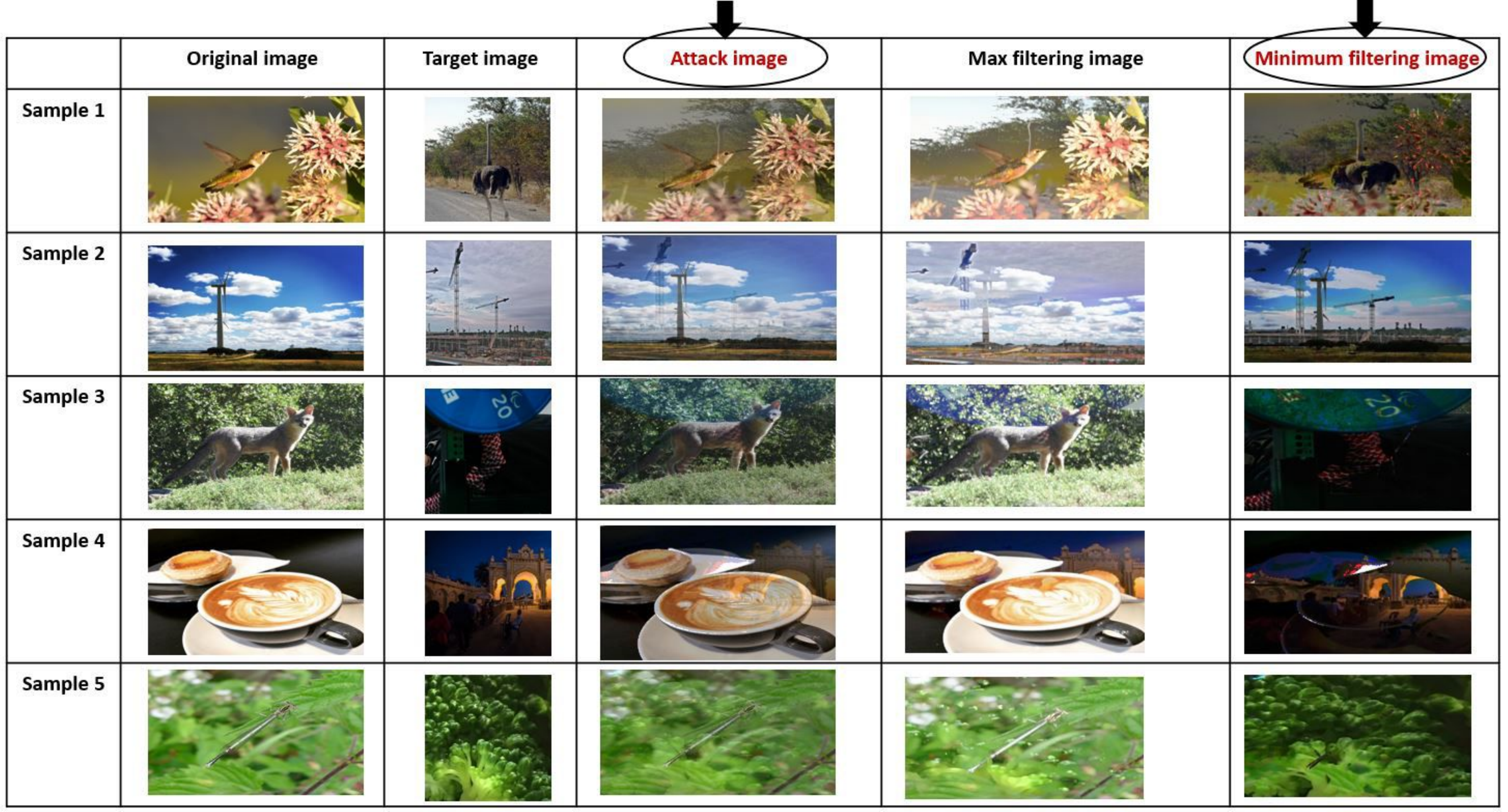}
\caption{More examples from our filtering detection method. The filtering mechanism especially the minimum filter consistently demonstrates an ability to reveal the embedded target image within the attack image.}
\label{fig:filtering_images}
\end{figure*}

Figure \ref{fig:filtering_images} presents visual examples to demonstrate the effectiveness of the proposed filtering detection method. We are able to quantify these results by using both MSE and SSIM metrics.

\section{Steganalysis detection method visual samples}
\label{append:Steganalysis detection method visual samples}

\begin{figure*}[!ht]
\centering
\includegraphics[width=0.9\linewidth]{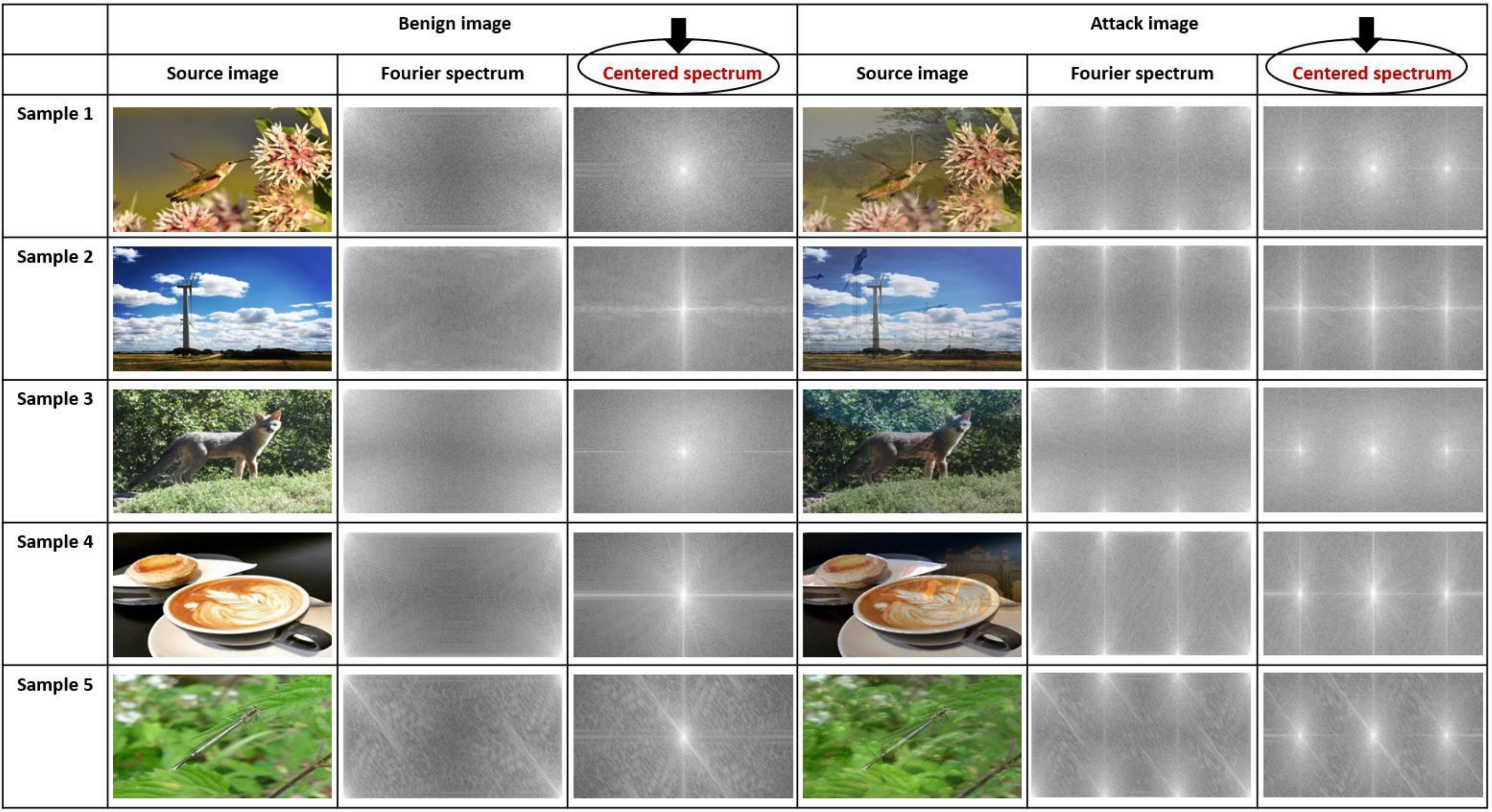}
\caption{More examples from our steganalysis detection method. We find image-scaling attack images consistently have more than \textit{three centered spectrum points} due to the abnormal perturbation of their pixels. On the other hand, the benign image has only \textit{one centered spectrum point}.}
\label{fig:steganalysis_images}
\end{figure*}

Figure \ref{fig:steganalysis_images} shows visual samples to exhibit the ability of our steganalysis method to detect the attack image by producing its centered spectrum points. 

